\documentclass[letterpaper]{article} 
\usepackage{aaai2026}  
\usepackage{times}  
\usepackage{helvet}  
\usepackage{courier}  
\usepackage[hyphens]{url}  
\usepackage{graphicx} 
\usepackage{natbib}  
\usepackage{caption} 
\usepackage{algorithm}

\usepackage{newfloat}
\usepackage{listings}
\DeclareCaptionStyle{ruled}{labelfont=normalfont,labelsep=colon,strut=off} 
\floatstyle{ruled}
\newfloat{listing}{tb}{lst}{}
\floatname{listing}{Listing}
\usepackage{subcaption}
\usepackage{booktabs}
\usepackage{tabularx}
\usepackage{multirow}
\usepackage{enumitem}
\usepackage{tipa}
\usepackage{amsmath}
\usepackage{algpseudocode} 

\newcommand{\dataset}{LEX-ICON}

\title{Do Language Models Associate Sound with Meaning? \\ A Multimodal Study of Sound Symbolism}
\author {
    Jinhong Jeong\equalcontrib\textsuperscript{\rm 1},
    Sunghyun Lee\equalcontrib\textsuperscript{\rm 1},
    Jaeyoung Lee\textsuperscript{\rm 2},
    Seonah Han\textsuperscript{\rm 3},
    Youngjae Yu\thanks{Corresponding author. J.J. introduced sound symbolism and performed the semantic dimension prediction. S.L. conducted the internal attention analysis. J.L. introduced the idea of formulating the problem as multimodal interpretability. S.H. built the constructed word data and interpreted the linguistic implications.}\textsuperscript{\rm 2}
}
\affiliations {
    \textsuperscript{\rm 1}Yonsei University\\
    \textsuperscript{\rm 2}Seoul National University\\
    \textsuperscript{\rm 3}Korea University\\
    \{jjhsnail0822, sheepswool\}@yonsei.ac.kr, \{jerry96, youngjaeyu\}@snu.ac.kr, sunahan@korea.ac.kr
}

\begin{document}

\maketitle

\begin{abstract}
\textit{Sound symbolism} is a linguistic concept that refers to non-arbitrary associations between phonetic forms and their meanings. We suggest that this can be a compelling probe into how Multimodal Large Language Models (MLLMs) interpret auditory information in human languages. We investigate MLLMs' performance on phonetic iconicity across textual (orthographic and IPA) and auditory forms of inputs with up to 25 semantic dimensions (e.g., \textit{sharp vs. round}), observing models' layer-wise information processing by measuring phoneme-level attention fraction scores. To this end, we present \textbf{\dataset}, an extensive mimetic word dataset consisting of 8,052 words from four natural languages (English, French, Japanese, and Korean) and 2,930 systematically constructed pseudo-words, annotated with semantic features applied across both text and audio modalities. Our key findings demonstrate (1) MLLMs' phonetic intuitions that align with existing linguistic research across multiple semantic dimensions and (2) phonosemantic attention patterns that highlight models' focus on iconic phonemes. These results bridge domains of artificial intelligence and cognitive linguistics, providing the first large-scale, quantitative analyses of phonetic iconicity in terms of MLLMs' interpretability.
\end{abstract}

\begin{links}
    \link{Code}{https://github.com/jjhsnail0822/sound-symbolism}
\end{links}

\section{Introduction}

\begin{figure}[tb]
    \centering
    \includegraphics[width=\columnwidth]{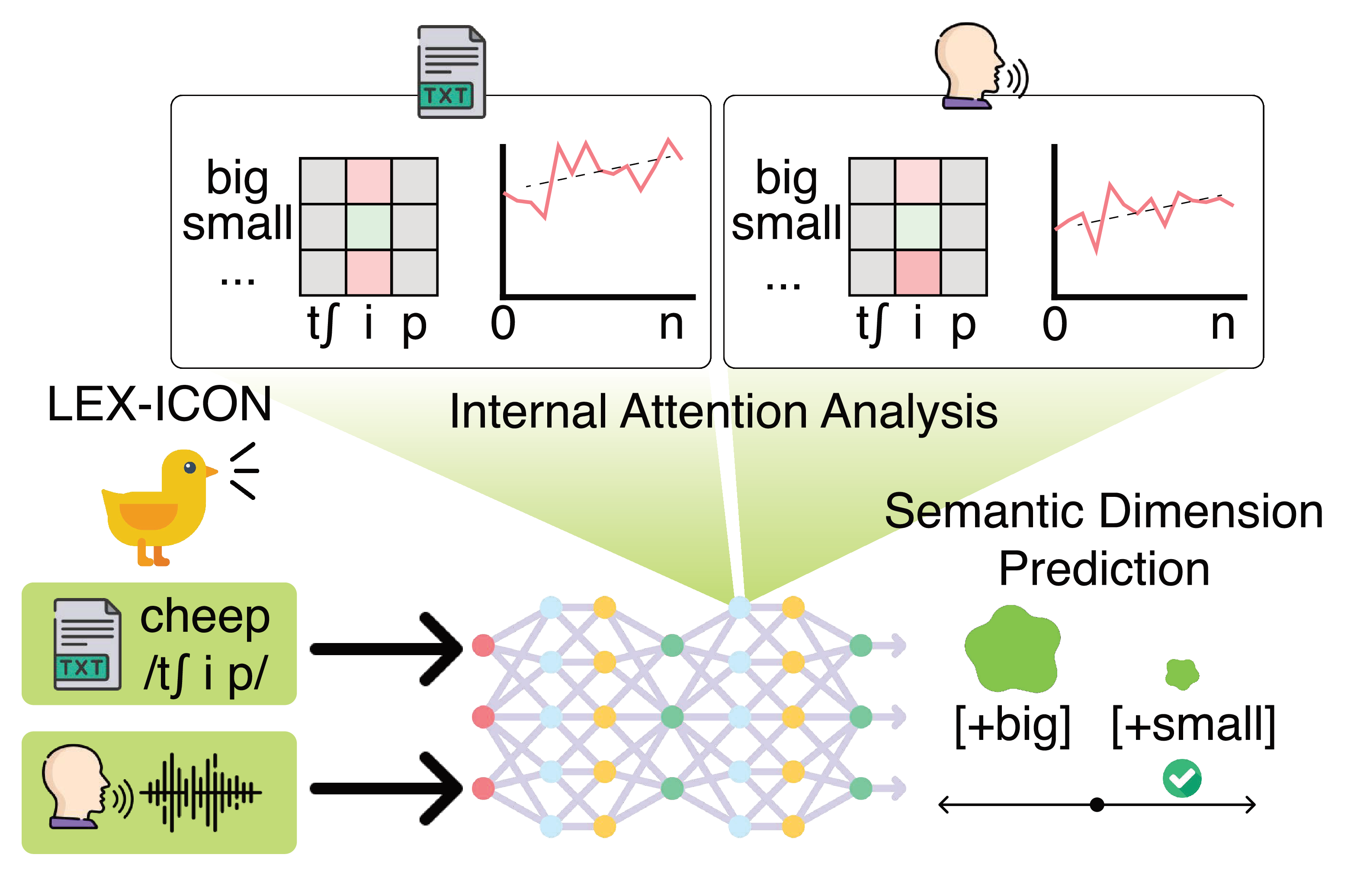}
    \caption{Phonetic iconicity investigation for MLLMs using natural and constructed mimetic words from text and audio modalities in \dataset. We conduct quantitative evaluations for up to 25 semantic dimensions and examine layer-wise attention fraction scores to identify how phonemes and meanings are related within the models.}
    \label{fig:figure_1}
\end{figure}

\textit{Sound symbolism}, which suggests that phonetic sounds and their meanings have a significant correlation, presents a cognitively grounded exception to the principle of linguistic arbitrariness~\citep{hinton2006sound, dingemanse2015arbitrariness}. For instance, when people are presented with two images of sharp and round shapes called ``kiki'' and ``bouba'', the overwhelming majority of participants consistently match ``kiki'' with pointed shapes and ``bouba'' with round shapes, regardless of their cultural background~\citep{kohler1967gestalt, ramachandran2001synaesthesia}. This iconicity effect illustrates the universal sensory associations of humans that intuitively link certain phonetic sounds with certain semantic features, which facilitate early-childhood or second-language acquisition~\citep{imai2014sound}, and commercially nuanced brand naming~\citep{yorkston2004sound}.

Recent advancements in Multimodal Large Language Models (MLLMs) that incorporate audio-modality input into an integrated representation space~\citep{openai2024gpt4ocard, xu2025qwen25omnitechnicalreport} shed light on new possibilities to systematically analyze human-like phonetic iconicity of language models. With MLLMs as a novel test bed, we formulate the following two key research questions:

\begin{enumerate}[label=\textit{RQ \arabic*.}, leftmargin=*, align=left]
    \item How do MLLMs associate sound-symbolic words with semantic features similar to human phonetic intuition?
    \item How do MLLMs' internal attention patterns align with phonosemantic relationships?
\end{enumerate}

\textit{Mimetic words} provide a compelling probe to address these subjects. Represented by onomatopoeias and ideophones\footnote{In this paper, ``onomatopoeia'' and ``ideophone'' refer to words that depict sounds and non-auditory sensory imagery, respectively.}, mimetic words refer to words in which the non-arbitrary association of sound form and meaning is prominent~\citep{akita2019ideophones}, such as ``boom'' or ``whizz.'' We construct \textbf{\dataset}, a dataset of natural mimetic words and constructed pseudo-words designed to maximize sound-symbolic effect. With \dataset, we apply the semantic dimension paradigm, where word meanings are projected onto binary scales (e.g., \textit{fast vs. slow}), to examine MLLMs' abilities to infer the meaning of words from their form (original text, IPA, audio) rather than their content. We further analyze models' internal representations to identify whether models actually focus on iconic phonemes, as illustrated in Figure~\ref{fig:figure_1}.

Our experimental results show that MLLMs demonstrate phonetic intuitions across multiple semantic dimensions, for not only natural words but also constructed pseudo-words that exclude models' memorization (e.g., \textit{sharp vs. round}). In particular, MLLMs exhibit modality-specific preferences across semantic dimensions, using audio for acoustically grounded features (e.g., \textit{big vs. small}) and text for articulatory or visually driven features (e.g., \textit{beautiful vs. ugly}). However, we reveal that there still remain discrepancies between the semantic dimensions where humans exhibit high scores and those where models perform well. Through phoneme-level attention analysis, we suggest that MLLMs attend to sound-symbolic phonemes corresponding to their meanings. In particular, we demonstrate that attention fraction to these sound-symbolic phonemes is more prominent in models' late-layers when processing constructed pseudo-words, suggesting that the relationship between phonemes and meanings can be systematically represented in the deep layers. Based on these findings, we summarize our key contributions as follows.

\begin{enumerate}
    \item We conduct the first large-scale investigation of phonetic iconicity in MLLMs, grounded in \dataset, a novel multilingual mimetic word dataset spanning multiple language families.
    \item We provide a quantitative evaluation methodology of sound symbolism through the semantic dimension approach that contributes to linguistics by capturing phonetic iconicity akin to humans.
    \item We present a comprehensive analysis that elucidates models' internal mechanisms towards sound symbolism by measuring phoneme-level attention scores, thereby contributing to the field of model interpretability.
\end{enumerate}

Our approach integrates two distinct but interconnected domains, addressing both artificial intelligence and linguistics. For the former, we observe an integration of form and meaning within MLLMs using sound symbolism probes, revealing new insights in terms of model interpretability. From a linguistic perspective, we propose a substantiation for nonhuman intelligence's phonetic intuitions on mimetic word data, providing a quantitative basis for linguistic experiments that have been conducted mostly with humans.

\section{Related Work}
\label{sec:related_work}

\subsection{Sound Symbolism in Linguistics}
Early modern linguists question the arbitrary relationship between signifier and signified, suggesting the phenomenon of intuitive phonetic symbolism in humans~\citep{usnadze1924experimenteller, sapir1929study, kohler1967gestalt}. \citet{sapir1929study}'s experiment has exhibited that people who are presented with unfamiliar object names ``mil'' and ``mal'' and are asked to estimate the size of the objects respond that the latter is larger than the former~\citep{parise2012audiovisual}. Recent studies continue to observe these phenomena~\citep{hinton2006sound, lockwood2015iconicity, cwiek2022bouba}, extending them by scaling to a large amount of data~\citep{thompson2021articulatory, winter2024iconicity} or measuring the exact semantic dimensions associated with each phoneme~\citep{monaghan2019sound, sidhu2022higher, sidhu2025sound}. For the mimetic words, experimental research has also demonstrated that people can infer a certain degree of meaning from the words in languages besides their mother tongue~\citep{shinohara2010cross, dingemanse2016sound}.

\subsection{Phonetic Iconicity for LLMs}

Iconicity tasks for language models have been used as a means of assessing whether the models have human-like phonetic intuition~\citep{cai-etal-2024-large, duan2024hlbbenchmarkingllmshumanlikeness}. Early studies focus on analyzing phonesthemic information contained in word embedding spaces~\citep{abramova2013automatic, abramova-fernandez-2016-questioning}. Recent studies demonstrate that the effect of non-arbitrary integration of linguistic form and meaning is substantiated in text-only LLMs~\citep{miyakawa-etal-2024-llms, marklova2025iconicitylargelanguagemodels}, vision and image generation models~\citep{loakman-etal-2024-ears, alper2024kikiboubasoundsymbolism, shinto2024analyzing, iida2024investigating}, and audio-visual models~\citep{tseng2024measuring}, yet they are not extensive enough in scope to include exhaustive and multilingual mimetic word data or diverse semantic feature analysis.

\subsection{Multimodal Interpretability}
Recently, the field of model interpretability has witnessed significant growth~\citep{elhage2021mathematical, zou2025representationengineeringtopdownapproach, wang2023interpretability}. Motivated by the rapid progress of MLLMs~\citep{openai2024gpt4ocard, xu2025qwen25omnitechnicalreport}, there has been increasing interest in multimodal interpretability~\citep{lin2025surveymechanisticinterpretabilitymultimodal}. For instance, \citet{neo2025towards} ablates a subset of visual tokens to observe the resulting differences in model output, while \citet{nikankin2025taskdifferentcircuitsdisentangling} demonstrates that models employ distinct processing circuits depending on the input modality. Despite these advances, prior work has predominantly focused on the visual modality, with less exploration of audio-based interpretability. \citet{yang2025audiolenscloserlookauditory} investigates how models handle auditory input, but their analysis is limited to simple settings where the audio simply serves as a direct vocalization of textual data. In this work, we examine audio interpretability through the lens of sound symbolism, highlighting how phoneme-level features encode meaning in ways that are intrinsically tied to the auditory modality.

\section{\dataset}
\label{sec:datasets}

\begin{figure*}[tb]
    \centering
    \includegraphics[width=\linewidth]{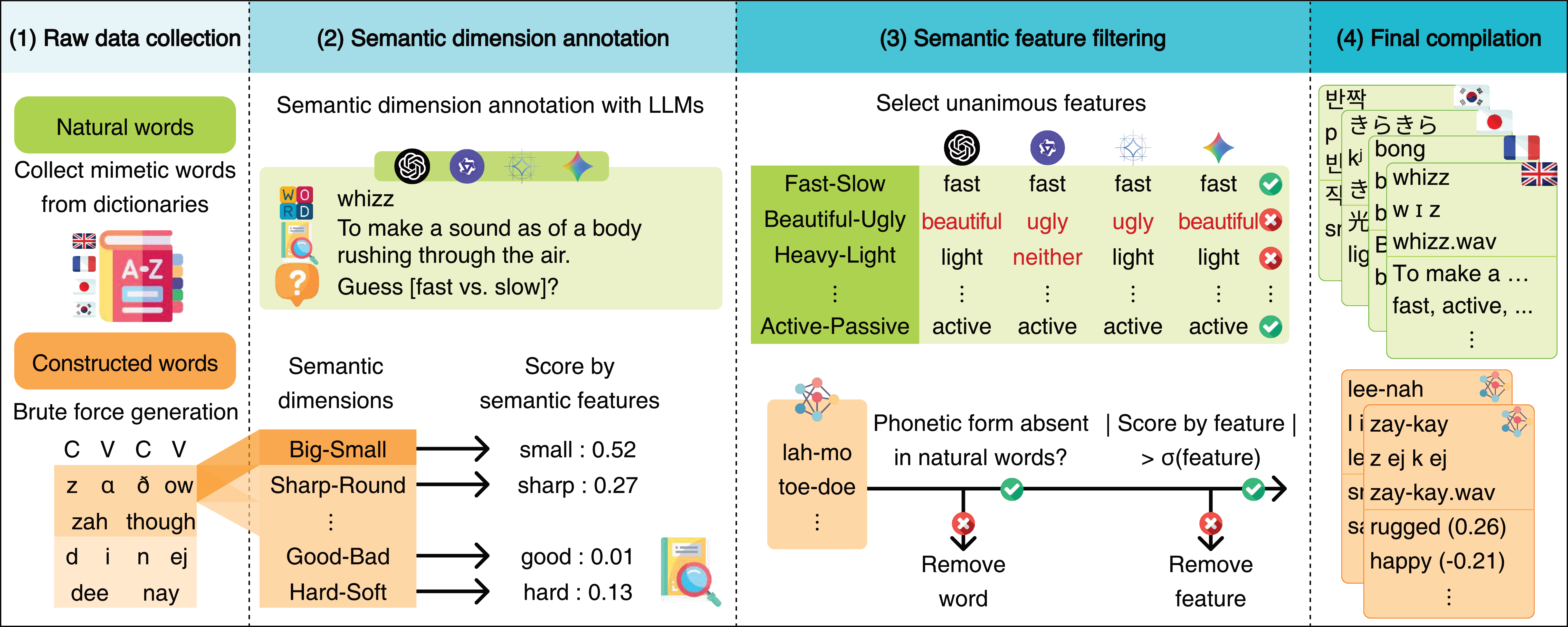}
    \caption{A comprehensive figure for the data construction flow of \dataset. (1) We manually collect 8,052 mimetic words and definitions from dictionaries, and systematically construct 2,930 disyllabic pseudo-words. (2) Using four LLMs (GPT-4.1, Qwen3-32B, Gemma-3-27B, and Gemini-2.5-flash), we automatically annotate each word with semantic dimensions based on its definitions. (3) For natural words, we retain features agreed upon by all models. For constructed words, we filter out features that are close to neutral. (4) The final dataset contains 10,982 words with 84,932 semantic features with varied input types.}
    \label{fig:data_construction_flow}
\end{figure*}

We build \dataset, a dataset with \textit{natural} word group consisting of existing large-scale mimetic words derived from four natural languages (English, French, Japanese, and Korean), as well as \textit{constructed} word group comprising systematically generated pseudo-words. Table~\ref{tab:dataset_distribution} and \ref{tab:semantic_dimension_pseudo_gt_simple_distribution} summarize the distribution of the overall datasets. Figure~\ref{fig:data_construction_flow} also illustrates an overall dataset construction flow.

\begin{table}[tb]
    \centering
        \begin{tabular}{lcccccc}
            \toprule
             & \textbf{En.} & \textbf{Fr.} & \textbf{Ja.} & \textbf{Ko.} & \textbf{Con.} & \textbf{Total} \\
            \midrule
            \# Words      & 826   & 809   & 1418  & 4999  & 2930 & \textbf{10982} \\
            \bottomrule
        \end{tabular}
    \caption{Data distribution of \dataset\ across \textit{natural} (8,052 words from English, French, Japanese, and Korean) and \textit{constructed} (2,930 words) groups. Japanese and Korean are known for their rich mimetic words~\citep{hamano1986sound, kwon2018iconicity}, which contributes to their high proportion.}
    \label{tab:dataset_distribution}
\end{table}

\begin{table}[htb]
\centering
\small
\begin{tabular}{lrrr}
\toprule
\textbf{Dimension} & \textbf{Natural} & \textbf{Constructed} & \textbf{Total} \\
\midrule
good-bad                  & 2083 & --    & 2083 \\
beautiful-ugly            & 929  & 462  & 1391 \\
pleasant-unpleasant       & 3380 & --    & 3380 \\
strong-weak               & 4299 & 208  & 4507 \\
big-small                 & 2073 & 1687 & 3760 \\
rugged-delicate           & 2664 & --    & 2664 \\
active-passive            & 3884 & --    & 3884 \\
fast-slow                 & 2051 & 1437 & 3488 \\
sharp-round               & 2323 & 1623 & 3946 \\
realistic-fantastical     & 6883 & 501  & 7384 \\
structured-disorganized   & 3712 & --    & 3712 \\
ordinary-unique           & 1585 & 208  & 1793 \\
interesting-uninteresting & 208  & --    & 208  \\
simple-complex            & 4322 & 1602 & 5924 \\
abrupt-continuous         & 4703 & 1005 & 5708 \\
exciting-calming          & 2256 & 1402 & 3658 \\
hard-soft                 & 3676 & 1136 & 4812 \\
happy-sad                 & 719  & 1463 & 2182 \\
harsh-mellow              & 3106 & 1005 & 4111 \\
heavy-light               & 2918 & 1341 & 4259 \\
inhibited-free            & 2673 & 206  & 2879 \\
masculine-feminine        & 378  & 1522 & 1900 \\
solid-nonsolid            & 2255 & 893  & 3148 \\
tense-relaxed             & 2956 & 206  & 3162 \\
dangerous-safe            & 448  & 541  & 989  \\
\midrule
\textbf{Total}            & \textbf{66484} & \textbf{18448} & \textbf{84932} \\
\bottomrule
\end{tabular}
\caption{Semantic dimension distribution of pseudo ground truth data by word group. We adopt 25 semantic dimension criteria by \citet{sidhu2022higher} to annotate \dataset. Six dimensions from the constructed word group are excluded by removing close-to-neutral data points. As a result, 19 dimensions remain for experiments in \S\ref{sec:semantic_dimension_prediction}.}
\label{tab:semantic_dimension_pseudo_gt_simple_distribution}
\end{table}

\subsection{Semantic Dimension}
\label{sec:semantic_dimension}

We employ the semantic dimension methodology, a concept originated from ``semantic differential'' by \citet{osgood1957measurement}, which consists of two semantic feature adjectives located at opposite extremes, such as ``big'' and ``small.'' By applying this methodology to sound symbolism experiments, we can simultaneously and precisely measure the multifactorial meanings contained in a single word. We annotate up to 25 pairs of predefined semantic features as per \citet{sidhu2022higher} for each word in the \dataset, which is the most diverse scale in terms of studies on LLMs' iconicity.

\subsection{Natural Mimetic Words}
\label{sec:multilingual_mimetic_words}

\paragraph{Data Collection.}

We manually collect 8,052 mimetic words and definition data consisting of onomatopoeias and ideophones in English, French, Japanese, and Korean from specialized mimetic word lexicons and authoritative dictionaries for each language. We then extract the most representative definitions of these mimetic words from dictionaries such as the Oxford English Dictionary~\citep{dictionary1989oxford}, Le Petit Robert~\citep{lepetitrobert2023}, Nihon Kokugo Daijiten~\citep{9784095210216}, and the Standard Korean Language Dictionary~\citep{nkl2025}. For further source information, see the Appendix.

\paragraph{Input Type Variation.}
To observe the effect of trained token memorization and modality change, we create three types of word form that contain textual and auditory input: (1) original text, (2) IPA-converted text with phoneme-level spacing, (3) audio waveform converted using text-to-speech (TTS) software. We perform IPA conversion with the Epitran package~\citep{Mortensen-et-al:2018}, and obtain the TTS dataset using Google Text-to-Speech~\citep{google_tts_api_2025} for English, French, and Japanese; and MeloTTS~\citep{zhao2024melo} for Korean.

\paragraph{Large-Scale Annotation Process.}
To effectively generate a large-scale ground truth data, all natural language words and their definitions in the \dataset\ are given four LLMs: GPT-4.1~\citep{openai2025_gpt41}, Qwen3-32B~\citep{yang2025qwen3technicalreport}\footnote{Non-reasoning mode.}, Gemma-3-27B~\citep{gemmateam2025gemma3technicalreport}, and Gemini-2.5-flash~\citep{comanici2025gemini25pushingfrontier}. The models are prompted to annotate each semantic dimension of a word with one of a feature pair or neutral labels (e.g., selecting one of the ``exciting'', ``calming'', or ``neither'' option) for each word. See the Appendix for the detailed prompt.

We finalize the results as pseudo ground truth by selecting unanimously agreed features across all models, deleting ``neither'' labels to remove the meaningless features for each word. As shown in Table~\ref{tab:semantic_dimension_pseudo_gt_simple_distribution}, this process filters 67.0\% of the total annotation points, yielding 66,484 high-quality semantic features for natural words. In \S\ref{sec:semantic_dimension_prediction}, we verify these pseudo ground truth data through a human evaluation experiment. For more information on the semantic dimension ground truth, refer to the Appendix.

\subsection{Constructed Mimetic Words}

\paragraph{Phoneme Combination Generation.}
We systematically construct novel words using a CVCV structure to create pseudo-words unlikely to be encountered during model training. We use 15 consonants from five categories: sonorants (/\textipa{l}/, /\textipa{m}/, /\textipa{n}/), voiced fricatives (/\textipa{v}/, /\textipa{D}/, /\textipa{z}/), voiceless fricatives (/\textipa{f}/, /\textipa{s}/, /\textipa{S}/), voiceless stops (/\textipa{p}/, /\textipa{t}/, /\textipa{k}/), and voiced stops (/\textipa{b}/, /\textipa{d}/, /\textipa{g}/). We use four vowels from two categories: front vowels (/\textipa{i}/, /\textipa{ej}/) and back vowels (/\textipa{A}/, /\textipa{ow}/). After filtering against existing entries in the IPA-dict database~\citep{ipa_dict} across four languages (US and UK English, French, Japanese, and Korean), we obtain 3,108 novel pseudo-words. We convert IPA symbols to English alphabet combinations for TTS compatibility and remove incorrectly pronounced words by Google TTS, yielding 2,930 final words.

\paragraph{Semantic Dimension Annotation.}
We utilize the empirical coefficients from \citet{sidhu2022higher}, who quantified associations between phoneme categories and semantic dimensions through human rating experiments. These coefficients represent how much phoneme categories deviate from the overall mean rating across 25 semantic dimensions. We assign corresponding scores to phonemes in our constructed words and calculate the mean score across the four phonemes. To focus on meaningful associations, we apply a threshold of 1.0 standard deviation from the neutral point, treating data points within this range and as semantically neutral, consistent with 'neither' labels used for multilingual mimetic words.

\section{Semantic Dimension Prediction}
\label{sec:semantic_dimension_prediction}

We perform semantic dimension A/B tests to answer \textit{RQ 1}, interpreting the results by semantic dimension, word group, and input type to quantitatively explore MLLMs' phonetic intuitions. Human evaluation is also conducted to ensure reliability of pseudo ground truth data in \dataset, and the results are further compared to those of MLLM experiments. Details of the experiments are provided in the Appendix.

\begin{figure*}[tb]
    \centering
    \includegraphics[width=\textwidth]{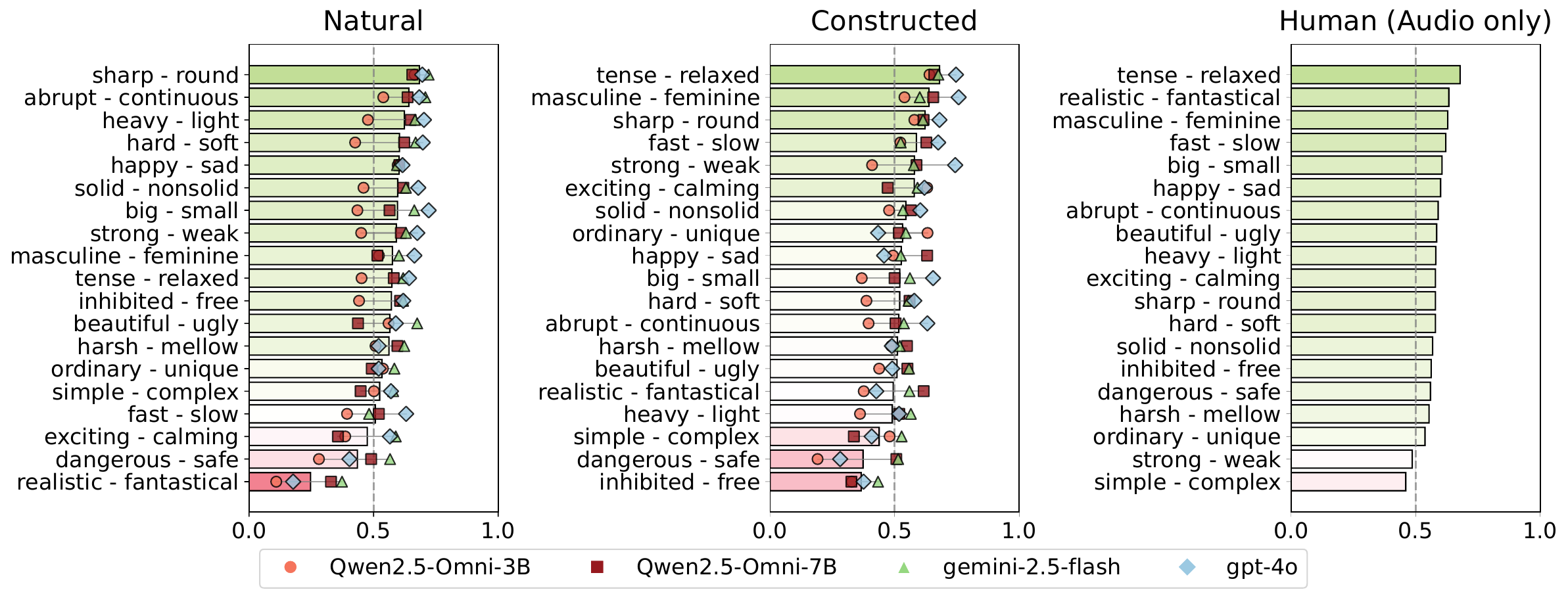}
    \caption{Macro-F1 score results for the semantic dimension A/B test. ``Natural'' and ``Constructed'' are results of LLM experiments, calculated by averaging all three input types (original text, IPA, and audio). Each dot represents each model's score for a given dimension. Human evaluation results only contain the ``Audio'' input type with sampled data for experimental feasibility, yet achieving superior scores compared to the baseline that demonstrate \dataset's reliability.}
    \label{fig:semantic_dimension_bar_comparison}
\end{figure*}

\subsection{Experimental Settings}

We employ MLLMs that officially support simultaneous text and audio inputs: Qwen2.5-Omni~\citep{xu2025qwen25omnitechnicalreport}\footnote{\texttt{Qwen2.5-Omni-3B} and \texttt{Qwen2.5-Omni-7B}.}, Gemini-2.5-flash~\citep{geminiteam2025geminifamilyhighlycapable}, and GPT-4o~\citep{openai2024gpt4ocard}\footnote{\texttt{gpt-4o} for text-only input, \texttt{gpt-4o-audio-preview} for audio-enabled input.}. All experimental models cover the four languages and orthographies present in \dataset. For the Qwen models, inference is performed on one RTX 4090 GPU. We apply a zero-shot prompting strategy with temperature set to 0 throughout all experiments to ensure reproducibility.

\subsection{Methodology}
\label{sec:semantic_dimension_methodology}

For each word, we prompt MLLMs with binary questions on each of the semantic dimensions and measure macro-F1 scores for the results of each dimension. We provide the models with words in three input types, keeping the query part as text. We insert audio tokens at the given words' positions within the prompt so that the audio tokens have the same series of positional embeddings with the text tokens, ensuring that the models infer internal representations in an integrated embedding space. An example prompt is illustrated in Table~\ref{tab:semantic_dimension_prompt_example}. The exact word input types are as follows:

\begin{itemize}
    \item Original text tokens (e.g., ``\textit{boom}'').
    \item Phoneme-level spaced IPA text tokens (e.g., ``\textit{b u m}'').
    \item TTS audio tokens (e.g., \texttt{<AUDIO>}\footnote{\texttt{<AUDIO>} represents a series of audio tokens that correspond to a given word.}).
\end{itemize}

\begin{table}[tb]
        \centering
        \begin{tabular}{p{0.9\linewidth}}
            \toprule
            \textbf{Semantic Dimension Test} \\
            \midrule
            Given an IPA [WORD] with its pronunciation audio, which semantic feature best describes the word based on auditory impression? \\
            \\
            \textnormal{[}WORD] \\
            \textbf{\textit{b u m}} (AUDIO: \texttt{<AUDIO>}) \\
            \\
            \textnormal{[}SEMANTIC DIMENSION] \\
            exciting vs. calming \\
            \midrule
            \textnormal{[}OPTIONS] \\
            \textbf{1: exciting} \\
            2: calming \\
            Answer with the number only. (1-2) \\
            \bottomrule
        \end{tabular}
    \caption{An example of prompts for the semantic dimension A/B test in \S\ref{sec:semantic_dimension_methodology}. This example illustrates a case of a natural language group English word ``boom'' that both the IPA text tokens and audio tokens (\texttt{<AUDIO>}) are presented. Detailed prompts are provided in the Appendix.}
    \label{tab:semantic_dimension_prompt_example}
\end{table}

We calculate macro-F1 scores to mitigate the imbalance in simple accuracy for each model across all combinations of word groups, input types, and semantic dimensions. Results for ``natural'' group are averaged equally across the four languages. Refer to the Appendix for detailed methods.

\subsection{Result}

\paragraph{Phonetic Intuition by Semantic Dimension.}

In Figure~\ref{fig:semantic_dimension_bar_comparison}, MLLMs' macro-F1 scores averaged across all input types surpass the baseline score (0.50) in 84.2\% (natural group) and 68.4\% (constructed group) of the semantic dimensions, indicating that the models can detect phonetic iconicity not only in natural mimetic words that may have been memorized during training phase, but also in constructed pseudo-words with maximized sound-symbolic effects. Overall performance becomes even larger when the comparatively small-scaled Qwen2.5-Omni-3B model is excluded. These results are supported by human evaluation results that score above the baseline in most dimensions, which guarantees the reliability of our pseudo ground truth data automatically annotated by LLMs from dictionary data in \S\ref{sec:semantic_dimension}. Notably, the models' strong performance on the \textit{sharp vs. round} dimension aligns with well-known cognitive linguistic experiments like the ``bouba-kiki'' effect~\citep{ramachandran2001synaesthesia}.

\paragraph{Human-like Iconicity.}

\begin{figure}[tb]
    \centering
    \includegraphics[width=\columnwidth]{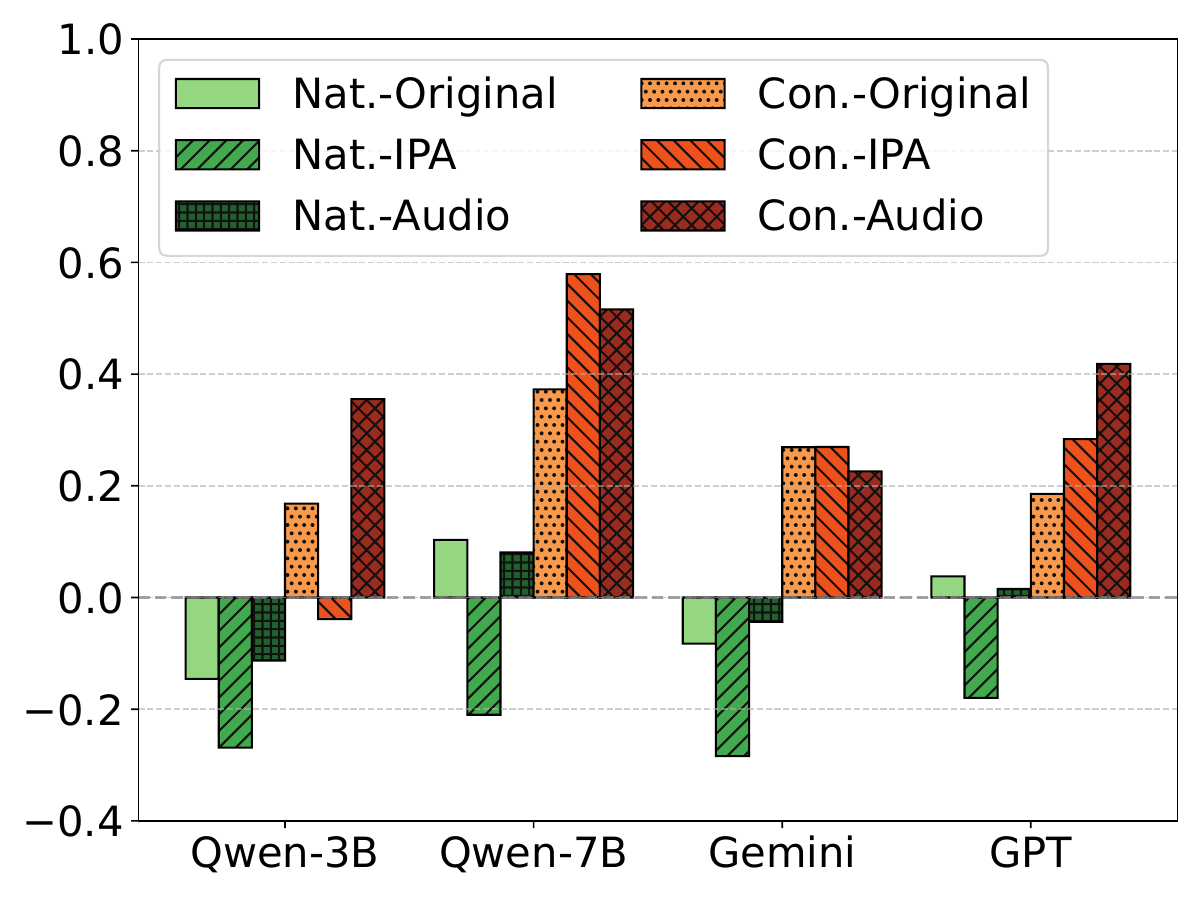}
    \caption{Pearson correlation scores with human evaluation results by word group and input type. Higher scores reflect greater similarity to humans' semantic dimension score distributions, where Qwen2.5-Omni-7B scores the highest correlation (maximum $r = 0.579$). In all models, constructed words elicit responses that are closer to human tendencies than natural words.}
    \label{fig:semantic_dimension_human_correlation_plot}
\end{figure}

Figure~\ref{fig:semantic_dimension_human_correlation_plot} shows that the Qwen2.5-Omni-7B model achieves the highest overall Pearson correlation coefficient with human evaluations across semantic dimensions, whereas larger models such as gemini-2.5-flash deviate more from human results. This suggests that while MLLMs can partially capture phonetic iconicity, they are still far from human-like semantic alignment. In particular, the relatively low correlation of IPA-converted natural word results indicates that linguistic arbitrariness from diverse languages may override iconic patterns even in mimetic words, as models' knowledge has been shaped by large-scale distributional semantics of natural languages, possibly leading to a tendency to overlook subtle phonosemantic cues.

\begin{figure}[tb]
    \centering
    \includegraphics[width=\columnwidth]{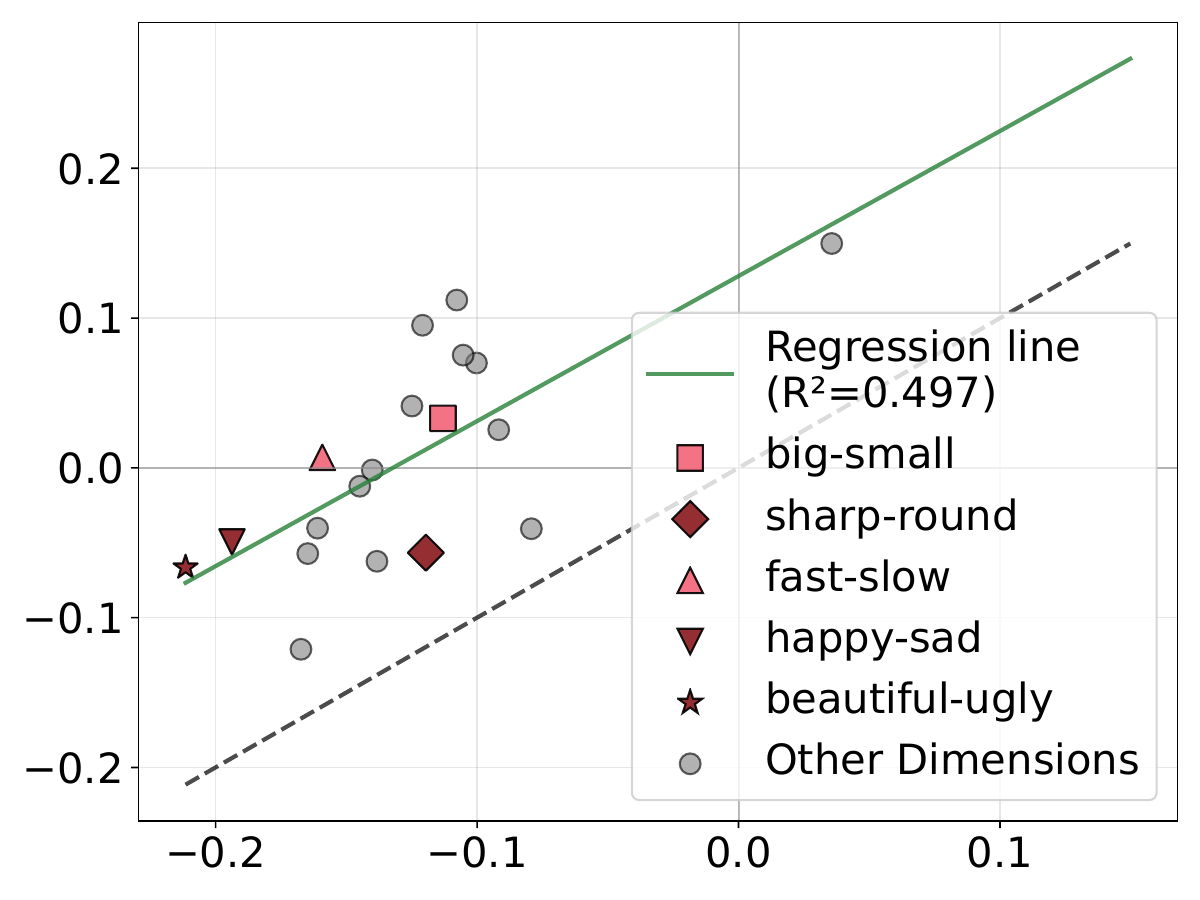}
    \caption{Advantage scores (macro-F1 differences) of audio inputs over the original text inputs by word group. X-axis indicates audio advantage scores for natural words, while Y-axis stands for constructed words. Each dot represents one semantic dimension, reflecting patterns aligned with linguistic implications, with an overall correlation (Pearson $r = 0.681$, Spearman $\rho = 0.705$).}
    \label{fig:modality_advantage_natural_constructed_audio_original}
\end{figure}

\paragraph{Linguistic Implication.}
 Our findings reveal systematic modality preferences across semantic dimensions, providing computational evidence for the multi-mechanism nature of sound symbolism, as shown in Figure~\ref{fig:modality_advantage_natural_constructed_audio_original}. For dimensions where acoustic features are theoretically central, such as size distinctions (\textit{big vs. small}) that correlate with formant frequencies \citep{knoeferle2017drives} and speed distinctions (\textit{fast vs. slow}) that relate to consonant voicing duration \citep{saji2013cross}, the MLLMs show a pattern of enhanced performance when processing constructed words in audio format, consistent with the theoretical predictions. Conversely, for dimensions where non-acoustic mechanisms are proposed, such as shape associations (\textit{sharp vs. round}) based on lip rounding gestures \citep{imai2025does} and valence associations (\textit{beautiful vs. ugly}, \textit{happy vs. sad}) affected by articulatory properties \citep{korner2022articulation}, the models exhibit stronger reliance on textural representations.

\section{Internal Attention Analysis}
\label{sec:internal_attention_analysis}
To address \textit{RQ 2}, internal attention analysis focuses on the internal phenomena that emerge during the MLLM inference process. Then these phenomena are compared with linguistic theories to evaluate their correspondence.

\begin{figure*}[tb]
    \centering
    \includegraphics[width=\textwidth]{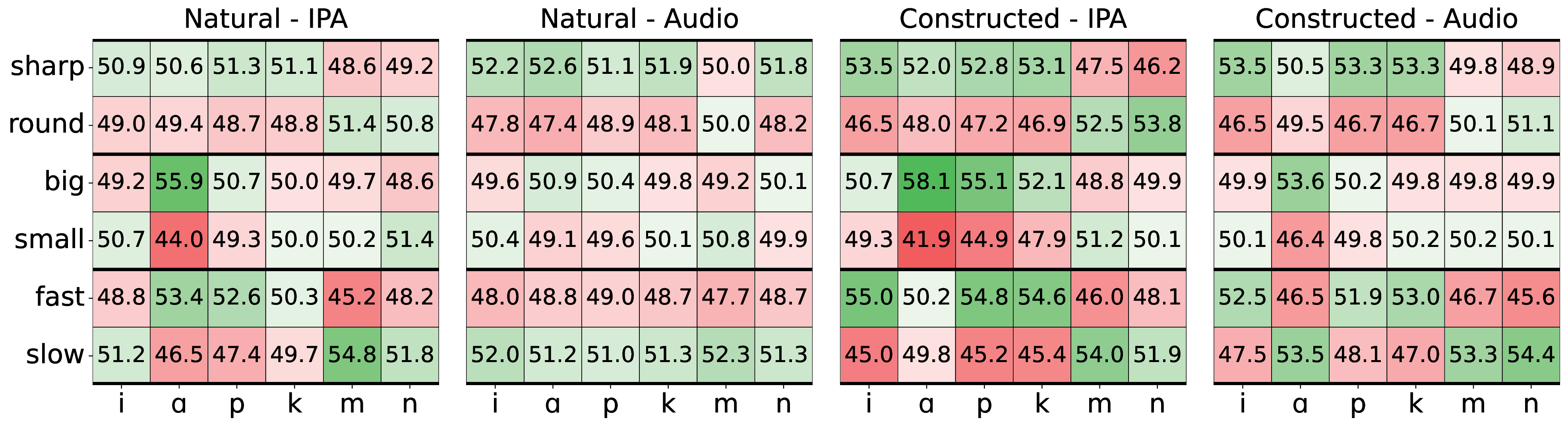}
    \caption{Attention fraction scores indicating the ratio of semantic dimensions that the model focuses on for each IPA, by word group and input type. For each IPA and semantic dimension, the model tends to attend more to semantic features that exhibit stronger associations with the given phonemes in linguistics. For instance, the model mostly associates \textit{sharp} semantic feature with /\textipa{p}/ and /\textipa{k}/, \textit{round} with /\textipa{m}/ and /\textipa{n}/, \textit{big} with /\textipa{A}/, and \textit{small} with /\textipa{i}/~\citep{kohler1967gestalt, parise2012audiovisual}.}
    \label{fig:attention_heatmaps}
\end{figure*}

\subsection{Experimental settings}
We utilize the Qwen2.5-Omni-7B model for the experiment due to its performance correlation most similar to humans, as well as its accessibility of model weights, which enables direct analysis of internal representations. All other settings for the experiment are as in \S\ref{sec:semantic_dimension_prediction} to maintain consistency.

\subsection{Methodology}
We calculate which semantic feature the model pays more attention to, for each IPA symbol given a semantic dimension. We perform layer-wise analyses of relative attention scores for each IPA symbol across contrasting semantic features, distinguishing by input types (IPA text and audio waveform) and word groups (natural and constructed words). Unlike original text tokens, the IPA text and audio input types facilitate phoneme-level investigation, as each phoneme is represented by at least one token.

\paragraph{Attention Fraction Score Extraction.}
For each inference conducted on the binary questions described in \S\ref{sec:semantic_dimension_prediction}, we obtain the attention scores only when the model generates the correct response. For each layer, we calculate attention scores between tokens corresponding to a single IPA symbol (e.g., /\textipa{w}/ from /\textipa{w} \textipa{I} \textipa{z}/) and tokens corresponding to each semantic feature (e.g., \textit{fast vs. slow}). After retrieving scores, we normalize the paired scores so that the sum of the two semantic feature scores is equal to one. These fraction scores undergo head-wise and word-wise averaging within each layer to yield a single mean score per layer. To mitigate the bias of assigning higher attention scores to preceding tokens, we repeat each experiment with the semantic features presented in reversed order and compute the mean normalized attention fraction score across both feature-order conditions.

\paragraph{Token-IPA Symbol Alignment.}
When an IPA symbol spans multiple text tokens, the procedure first sums the attention scores across those tokens before fraction normalization. For audio inputs, we employ Montreal Forced Aligner~\citep{mcauliffe2025montreal} to segment each audio waveform into phonemes over time. The phoneme sequence is then aligned with the model’s 40 ms sampling period, with consecutive occurrences of the same phoneme treated as a single IPA symbol. Further details are in the Appendix.

\subsection{Result}
An attention fraction score above 0.5 for a given IPA-semantic feature pair indicates the model's preferential focus on phonemes with sound-symbolic associations. For more details about the experiment, refer to the Appendix.

\begin{figure}[tb]
    \centering
    \includegraphics[width=\columnwidth]{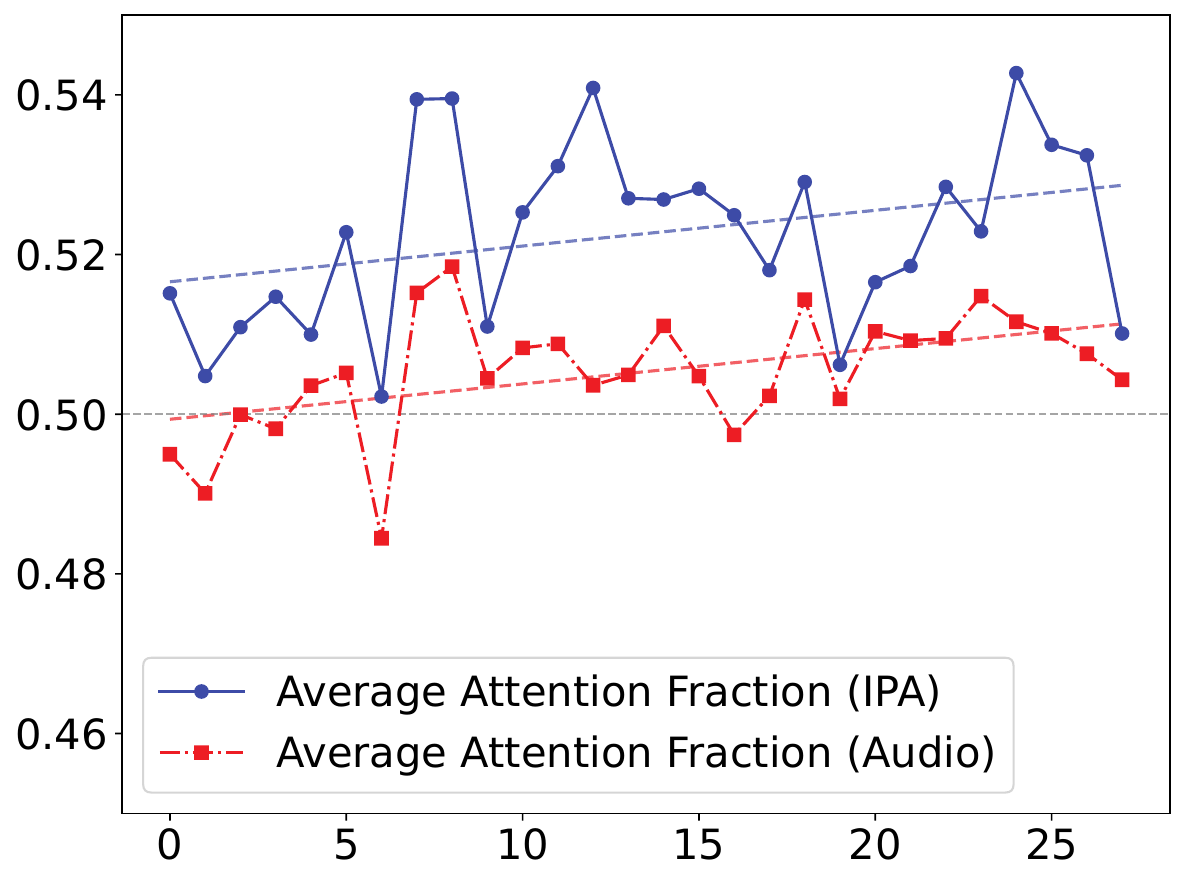}
    \caption{Attention fraction scores for the constructed word group, averaged across semantic dimensions. X-axis indicates the model's attention layer number, where Y-axis means the ratio at which the model attend to the correct semantic feature for a phoneme. While most layers score above the baseline fraction (0.50), the model focus on IPA text more than audio input for given iconic phonemes. Furthermore, the model tend to concentrate more on iconic phonemes in its late-layers.}
    \label{fig:layer_attention_score_plot}
\end{figure}

\paragraph{Layer-wise Attention Fraction Score.}
Figure~\ref{fig:layer_attention_score_plot} demonstrates that, for constructed words, the attention fraction scores for IPA text consistently exceed those for audio input type across layers (IPA = 0.523, audio = 0.506 on average), showing an upward trend toward the late layers. These lower phoneme‐level attention scores on audio inputs may imply that multimodal models derive greater benefits from extensively trained texts than from the acoustic properties of less-trained audio data. On the other hand, the average attention fraction scores for natural words (IPA = 0.507, audio = 0.501 on average) are lower than those for constructed words. This phenomenon may occur because arbitrary form–meaning mappings of natural words attenuate phonosemantic cues, thereby obscuring iconic phonemes. This interpretation is further corroborated by Figure~\ref{fig:semantic_dimension_human_correlation_plot}, which shows the low correlation between human evaluation scores for natural words in both IPA text and audio modalities.

\paragraph{Phoneme-Semantic Feature Relation.}
Figure~\ref{fig:attention_heatmaps} presents heatmaps of attention fraction scores for canonical IPA symbols and semantic dimensions by input type and word group. The patterns for constructed words in the IPA modality closely mirror prior findings in sound symbolism research. For example, phonemes such as /\textipa{p}/ and /\textipa{k}/ exhibit elevated attention under the \textit{sharp} feature, whereas /\textipa{m}/ and /\textipa{n}/ associate with the \textit{round} feature~\citep{kohler1967gestalt}.

\section{Conclusion}

In this work, we investigate MLLMs' phonetic iconicity on natural and constructed words via semantic dimension and internal attention analysis, constructing \dataset, a large-scale mimetic word dataset for MLLM analysis for the first time. We discover that MLLMs have the ability to detect sound symbolism in both natural and constructed mimetic words, and pay a higher rate of attention in the internal layers to iconic phonemes. These results suggest that the models' sound-meaning association can be explained in terms of interpretability. Future work could further deepen the analytical methodology presented in this work through experiments with more human participants, investigate modality-specific information such as intonation, or extend it to application fields such as language learning or brand effects.

\section*{Acknowledgments}

This work was partly supported by an Institute of Information \& communications Technology Planning \& Evaluation (IITP) grant funded by the Korean
Government (MSIT) (No.~RS-2021-II211343, Artificial Intelligence Graduate School Program (Seoul National University), No.RS-2025-02263598, Development of Self-Evolving Embodied AGI Platform Technology through Real-World Experience), the National Research Foundation of Korea(NRF) grant funded by the Korea government(MSIT)(RS-2024-00354218, RS-2024-00353125). We express special thanks to KAIT GPU project. The ICT at Seoul National University provides research facilities for this study.

\bibliography{aaai2026}

\appendix
\label{sec:appendix}

\section{Details on \dataset}
\label{appendix:detail_dataset}

\subsection{Mimetic Word Data}

\paragraph{Natural Word Collection.}
We manually extract headwords and definitions from published onomatopoeia dictionaries for each language to determine whether a word belongs to the mimetic word category. For Korean, the word definition guideline of the \citet{NIKL_StandardKoreanLanguageDictionaryCompilationGuideline2_2000} is used to extract mimetic words and definitions. Semantic definitions that need to be supplemented are obtained from authoritative dictionaries for each language. The full list of the dictionaries is provided in Table~\ref{tab:language_resources}.

\begin{table*}[htb]
\centering
\begin{tabular}{{p{0.1\textwidth}p{0.8\textwidth}}}
\toprule
\textbf{Language} & \textbf{References} \\
\midrule
English & Rhyme over Reason: Phonological Motivation in English~\citep{benczes2019rhyme} \\
        & Oxford English Dictionary, 2nd Edition~\citep{dictionary1989oxford} \\
\addlinespace
French  & Dictionnaire des Onomatop{\'e}es~\citep{enckell2003dictionnaire} \\
        & Le Petit Robert de la Langue Française~\citep{lepetitrobert2023} \\
        & Trésor de la Langue Française~\citep{letresor1994} \\
\addlinespace
Japanese & Kurashi no Kotoba Gion Gitaigo Jiten~\citep{1130282271732203008} \\
         & Seisenban Nihon Kokugo Daijiten~\citep{9784095210216} \\
         & Kojien, 7th Edition~\citep{kojien7} \\
         & Digital Daijisen~\citep{digitaldaijisen} \\
         & Daijirin, 4th Edition~\citep{matsumura2019daijirin} \\
\addlinespace
Korean & Standard Korean Language Dictionary~\citep{nkl2025} \\
\bottomrule
\end{tabular}
\caption{Reference materials for natural mimetic word collection by language.}
\label{tab:language_resources}
\end{table*}

We filter the data obtained from the mimetic word dictionaries, consisting of 954 English words, 1049 French words, 2025 Japanese words, and 4999 Korean words, using information from the aforementioned authoritative dictionaries. We finally determine 826 English words, 809 French words, 1418 Japanese words, and 4999 Korean words as natural mimetic word data.

\paragraph{Constructed Word Generation.}
We apply systematic pre-processing to IPA-dict entries from open-dict-data dictionaries during the filtering process to exclude words that have the same pronunciation as natural language words. We remove non-distinctive markers including stress (\textipa{"}) and duration (\textipa{:}) symbols, and convert allophones to their corresponding phonemes such as changing velarized lateral [\textltilde] to /\textipa{l}/ in English and tap [\textipa{R}] to /\textipa{l}/ in Korean. We also standardize notation by converting /\textipa{eI}/ to /\textipa{ej}/, /\textipa{oU}/ to  /\textipa{ow}/.

For TTS compatibility, we create IPA-to-alphabet mapping rules and refine mappings where needed, which includes converting ``tho'' to ``though'' to ensure correct /\textipa{Dow}/ pronunciation. Generated words are hyphenated (e.g., ``lah-mo'') except for combinations ending in /\textipa{Di}/ (e.g., ``laythey''). 

\paragraph{Word Group.}

\begin{figure}[htb]
    \centering
    \includegraphics[width=\columnwidth]{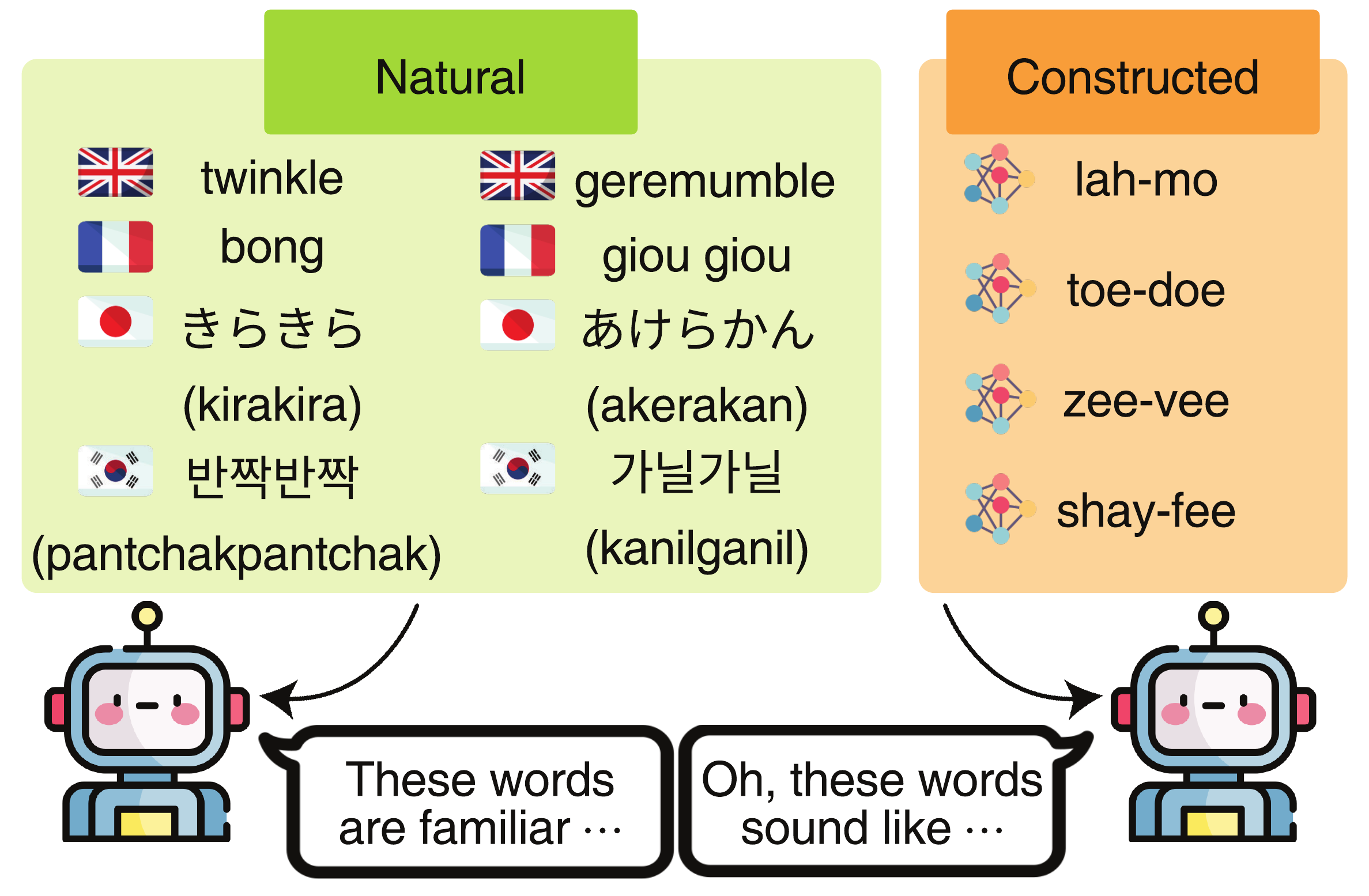}
    \caption{An illustrative description of the word groups. \textit{Natural} word group consists of English, French, Japanese, and Korean mimetic words. \textit{Constructed} word group includes disyllabic pseudo-words that experimental models may not have trained.}
    \label{fig:word_group_description}
\end{figure}

In Figure~\ref{fig:word_group_description}, we illustrate example words of the \textit{natural} and \textit{constructed} word groups that comprise \dataset. 

\paragraph{Input Type.}

\begin{figure}[htb]
    \centering
    \includegraphics[width=\columnwidth]{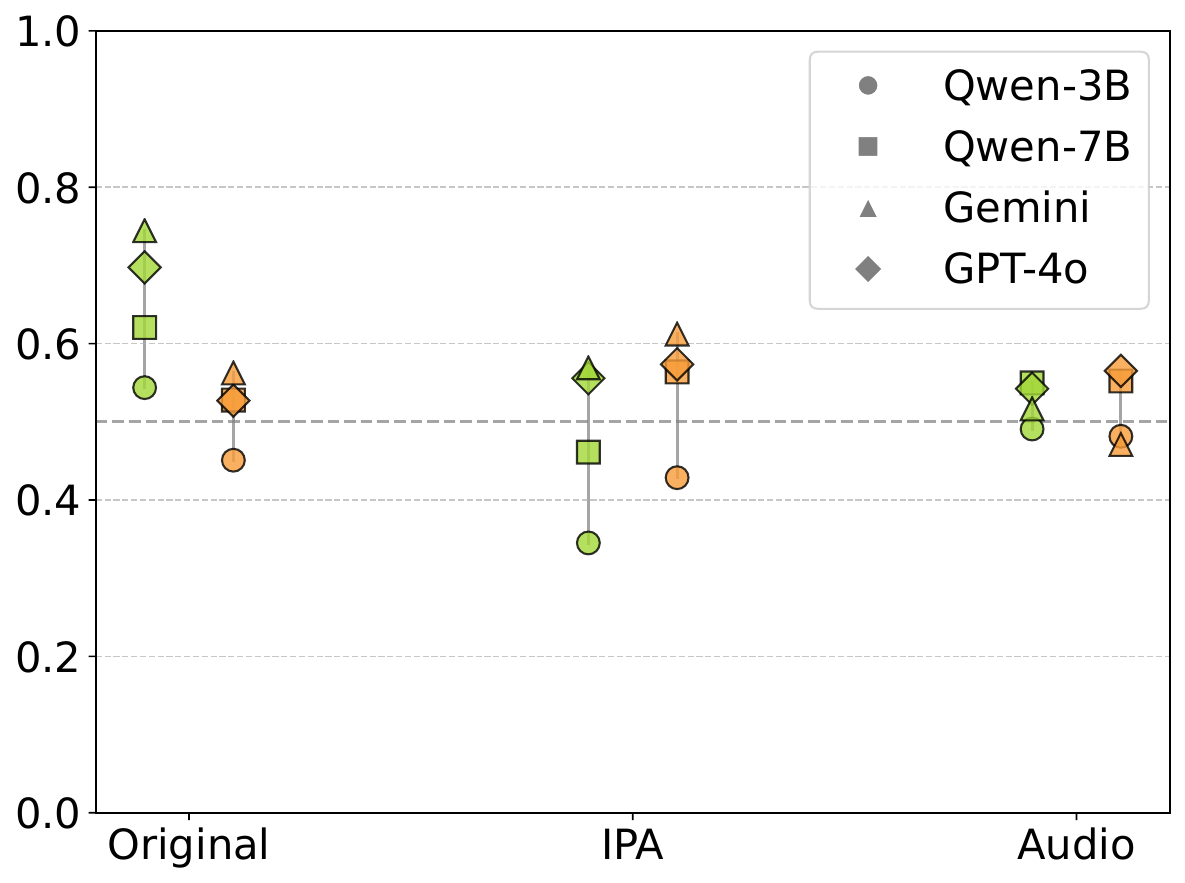}
    \caption{Semantic dimension prediction performance of models. For each input type, the dots for natural words' results are on the left, while constructed words' results are on the right. Each dot indicates macro-F1 scores of the models' response averaged across semantic dimensions.}
    \label{fig:semantic_dimension_input_type_scatter}
\end{figure}

We convert each word into three types (original orthographic text, phoneme-level spaced IPA text, and text-to-speech audio waveform) and input them into the experimental models. Figure~\ref{fig:semantic_dimension_input_type_scatter} shows the semantic dimension prediction performance of models for each word group and input type.

\subsection{Semantic Dimension}
Table~\ref{tab:semantic_dimension_list} shows the extreme feature pairs of 25 semantic dimensions suggested by \citet{sidhu2022higher}, which have been utilized for our semantic dimension prediction experiments.

After the automatic annotation process for natural words is complete, we utilize labels that are agreed upon by all model families and not annotated as ``neither'' as pseudo ground truth. Similarly, we remove semantic features close to neutral from the dataset by deleting semantic dimensions from each constructed word whose semantic features' ground truth score is within 1.0 standard deviation of the overall distribution. Due to this method, words in the datasets have fewer than 25 dimension labels. The distribution of these labels is shown in Table~\ref{tab:semantic_dimension_pseudo_gt_distribution}.

\begin{table*}[ht]
\centering
\small
\begin{tabular}{lrrrrrrrrr}
\toprule
\textbf{Dimension} & \multicolumn{3}{c}{\textbf{Natural}} & \multicolumn{3}{c}{\textbf{Constructed}} & \multicolumn{3}{c}{\textbf{Total}} \\
\cmidrule(lr){2-4} \cmidrule(lr){5-7} \cmidrule(lr){8-10}
(Feature 1)-(Feature 2) & Ft. 1 & Ft. 2 & Total & Ft. 1 & Ft. 2 & Total & Ft. 1 & Ft. 2 & Total \\
\midrule
good-bad                  & 532  & 1551 & 2083 & 0   & 0   & 0   & 532  & 1551 & 2083 \\
beautiful-ugly            & 290  & 639  & 929  & 344 & 118 & 462 & 634  & 757  & 1391 \\
pleasant-unpleasant       & 797  & 2583 & 3380 & 0   & 0   & 0   & 797  & 2583 & 3380 \\
strong-weak               & 2273 & 2026 & 4299 & 107 & 101 & 208 & 2380 & 2127 & 4507 \\
big-small                 & 902  & 1171 & 2073 & 894 & 793 & 1687 & 1796 & 1964 & 3760 \\
rugged-delicate           & 1327 & 1337 & 2664 & 0   & 0   & 0   & 1327 & 1337 & 2664 \\
active-passive            & 3267 & 617  & 3884 & 0   & 0   & 0   & 3267 & 617  & 3884 \\
fast-slow                 & 1579 & 472  & 2051 & 750 & 687 & 1437 & 2329 & 1159 & 3488 \\
sharp-round               & 1644 & 679  & 2323 & 789 & 834 & 1623 & 2433 & 1513 & 3946 \\
realistic-fantastical     & 6876 & 7    & 6883 & 251 & 250 & 501 & 7127 & 257  & 7384 \\
structured-disorganized   & 705  & 3007 & 3712 & 0   & 0   & 0   & 705  & 3007 & 3712 \\
ordinary-unique           & 229  & 1356 & 1585 & 101 & 107 & 208 & 330  & 1463 & 1793 \\
interesting-uninteresting & 135  & 73   & 208  & 0   & 0   & 0   & 135  & 73   & 208  \\
simple-complex            & 2064 & 2258 & 4322 & 784 & 818 & 1602 & 2848 & 3076 & 5924 \\
abrupt-continuous         & 2055 & 2648 & 4703 & 440 & 565 & 1005 & 2495 & 3213 & 5708 \\
exciting-calming          & 1741 & 515  & 2256 & 693 & 709 & 1402 & 2434 & 1224 & 3658 \\
hard-soft                 & 1757 & 1919 & 3676 & 440 & 696 & 1136 & 2197 & 2615 & 4812 \\
happy-sad                 & 436  & 283  & 719  & 709 & 754 & 1463 & 1145 & 1037 & 2182 \\
harsh-mellow              & 2477 & 629  & 3106 & 440 & 565 & 1005 & 2917 & 1194 & 4111 \\
heavy-light               & 1048 & 1870 & 2918 & 622 & 719 & 1341 & 1670 & 2589 & 4259 \\
inhibited-free            & 1484 & 1189 & 2673 & 105 & 101 & 206 & 1589 & 1290 & 2879 \\
masculine-feminine        & 144  & 234  & 378  & 689 & 833 & 1522 & 833  & 1067 & 1900 \\
solid-nonsolid            & 933  & 1322 & 2255 & 105 & 788 & 893  & 1038 & 2110 & 3148 \\
tense-relaxed             & 1986 & 970  & 2956 & 105 & 101 & 206  & 2091 & 1071 & 3162 \\
dangerous-safe            & 392  & 56   & 448  & 440 & 101 & 541  & 832  & 157  & 989  \\
\bottomrule
\end{tabular}
\caption{Semantic dimension distribution of pseudo ground truth by word group.}
\label{tab:semantic_dimension_pseudo_gt_distribution}
\end{table*}

\begin{table}[htb]
\centering
    \begin{tabularx}{0.9\linewidth}{@{}X X@{}}
        \toprule
            \textbf{Feature A} & \textbf{Feature B} \\
            \midrule
            good & bad \\
            beautiful & ugly \\
            pleasant & unpleasant \\
            \midrule
            strong & weak \\
            big & small \\
            rugged & delicate \\
            \midrule
            active & passive \\
            fast & slow \\
            sharp & round \\
            \midrule
            realistic & fantastical \\
            structured & disorganized \\
            ordinary & unique \\
            interesting & uninteresting \\
            simple & complex \\
            \midrule
            abrupt & continuous \\
            exciting & calming \\
            hard & soft \\
            happy & sad \\
            harsh & mellow \\
            heavy & light \\
            inhibited & free \\
            masculine & feminine \\
            solid & nonsolid \\
            tense & relaxed \\
            \midrule
            dangerous & safe \\
        \bottomrule
    \end{tabularx}
\caption{The 25 semantic dimension list by \citet{sidhu2022higher}.}
\label{tab:semantic_dimension_list}
\end{table}

\subsection{An Example Entry from \dataset}

\dataset\ consists of the following information for each mimetic word, as shown in Table~\ref{tab:dataset_example_entry}.

\begin{table}[htb]
\centering
\begin{subtable}[t]{0.9\linewidth}
\centering
\begin{tabular}{p{0.3\linewidth} p{0.6\linewidth}}
\toprule
\textbf{Key} & \textbf{Value} \\
\midrule
\texttt{word} & whizz \\
\texttt{meaning} & To make a sound as of a body rushing through the air. \\
\texttt{ref} & Rhyme Over Reason: Phonological Motivation in English \\
\texttt{ipa} & \textipa{w I z} \\
\texttt{romanization} & whizz \\
\texttt{language} & en \\
\bottomrule
\end{tabular}
\caption{Metadata from an example word.}
\label{tab:dataset_example_entry_metadata}
\end{subtable}

\vspace{0.5cm} 

\begin{subtable}[t]{0.9\linewidth}
\centering
\begin{tabular}{p{0.6\linewidth} p{0.3\linewidth}}
\toprule
\textbf{Dimension} & \textbf{Feature} \\
\midrule
\texttt{active-passive} & active \\
\texttt{fast-slow} & fast \\
\texttt{sharp-round} & sharp \\
\texttt{realistic-fantastical} & realistic \\
\texttt{simple-complex} & simple \\
\texttt{abrupt-continuous} & continuous \\
\texttt{exciting-calming} & exciting \\
\texttt{harsh-mellow} & harsh \\
\texttt{inhibited-free} & free \\
\texttt{solid-nonsolid} & nonsolid \\
\bottomrule
\end{tabular}
\caption{Semantic dimension data from an example word.}
\label{tab:dataset_example_entry_dimensions}
\end{subtable}
\caption{An example entry from \dataset.}
\label{tab:dataset_example_entry}
\end{table}

\section{Experimental Settings}

\subsection{Inference Settings}

\begin{itemize}
    \item Hardware: four NVIDIA RTX 4090 GPUs for 27B and 32B models, one NVIDIA RTX 4090 GPU for 3B and 7B models, and API calls for other proprietary models.
    \item Software: PyTorch~\citep{paszke2017automatic} and vLLM~\citep{kwon2023efficient} libraries.
    \item Random seed: 42 (for reproducibility, only if needed).
    \item Temperature: 0 (for reproducibility).
    \item Maximum output tokens: 1024.
\end{itemize}

\subsection{Language Models Used in This Work}

In this work, we utilize the following language models for data annotation, experimentation, and analysis.

\begin{itemize}
    \item gpt-4o~\citep{openai2024gpt4ocard}
    \item gpt-4o-audio-preview~\citep{openai2024gpt4ocard}
    \item gpt-4.1~\citep{openai2025_gpt41}
    \item gemini-2.5-flash~\citep{comanici2025gemini25pushingfrontier}
    \item gemma-3-27b-it~\citep{gemmateam2025gemma3technicalreport}
    \item Qwen3-32B~\citep{yang2025qwen3technicalreport}
    \item Qwen2.5-Omni-3B~\citep{xu2025qwen25omnitechnicalreport}
    \item Qwen2.5-Omni-7B~\citep{xu2025qwen25omnitechnicalreport}
\end{itemize}

\section{Token-IPA Symbol Alignment}
To enable phoneme-level attention analysis, we implement a pipeline for aligning audio tokens with IPA symbols.

\subsection{Montreal Forced Alignment (MFA) Pipeline}

We employ Montreal Forced Alignment \citep{mcauliffe2025montreal} to obtain precise phoneme-level segmentation of audio waveforms. We utilize pre-trained acoustic models for each language:
\begin{itemize}
    \item English (US) ARPA acoustic model v3.0.0  \citep{mfa_english_us_arpa_acoustic_2024}
    \item French MFA acoustic model v3.0.0~\citep{mfa_french_mfa_acoustic_2024}
    \item Japanese MFA acoustic model v3.0.0~\citep{mfa_japanese_mfa_acoustic_2024}
    \item Korean MFA acoustic model v3.0.0~\citep{mfa_korean_mfa_acoustic_2024}
\end{itemize}

The alignment process generates Textgrid files containing phoneme boundaries with millisecond precision.

\subsection{Custom Dictionary Generation and OOV Handling}

We generate language-specific pronunciation dictionaries by mapping word-IPA pairs from our dataset to MFA phoneme representations. When initial alignment fails due to out-of-vocabulary (OOV) words, we automatically
(1) identify words producing `spn' (spoken noise) tokens in TextGrid output,
(2) generate pronunciations using Epitran for OOV words,
(3) update pronunciation dictionaries with new entries, and (4) re-run MFA alignment until convergence.

\subsection{TextGrid Parsing and Audio Token Alignment}
We parse MFA-generated TextGrid files to extract phoneme boundaries and convert them into discrete frame sequences compatible with the model's 40ms sampling period. For each audio frame, we assign the phoneme whose temporal boundary encompasses the frame's center point. The resulting frame-level sequence is then aligned with the model's audio token representations, enabling precise measurement of phoneme-level attention patterns for audio inputs. 

\section{Metrics}
\label{appendix:metrics}

\subsection{Semantic Dimension Prediction}

\paragraph{Semantic Dimension Macro-F1 Score.}

\begin{itemize}
    \item $S$: Set of semantic dimensions, $|S|=d$.
    \item $C$: Set of semantic features in each semantic dimension. $|C|=2$.
    \item $TP_{s,c}$: True–positive count for dimension $s$ and feature $c$.
    \item $FP_{s,c}$: False–positive count for dimension $s$ and feature $c$.
    \item $FN_{s,c}$: False–negative count for dimension $s$ and feature $c$.
\end{itemize}

The macro-F1 score for each semantic dimension $s$ is defined as follows:

{
\[
\mathrm{MacroF1}_{s} \;=\;
\frac{1}{|C|}
\sum_{c \in C}
\frac{2\,TP_{s,c}}
     {2\,TP_{s,c} + FP_{s,c} + FN_{s,c}}
\]
}

We calculate the final natural group scores for each semantic dimension of LLMs by averaging the macro-F1 scores for four natural languages (English, French, Japanese, and Korean) and three input types (original text, IPA text, and audio waveform) for each dimension. For the constructed words, only averaging three input types is performed with the same procedure as in the calculation of natural words' scores.

\paragraph{Correlation Score with Human Evaluation.}

\begin{itemize}
    \item $D_{\text{valid}}$: the set of semantic dimensions for which both human evaluation macro-F1 scores and a model's macro-F1 scores are available.
    \item $h_d$: the human evaluation macro-F1 score for dimension $d$ and audio input type.
    \item $m_d$: the model's macro-F1 score for dimension $d$ and a given input type.
    \item $\mu_H$: the mean of the human scores across all dimensions in $D_{\text{valid}}$.
    \item $\mu_M$: the mean of the model's scores across all dimensions in $D_{\text{valid}}$.
\end{itemize}

The Pearson correlation coefficient $r$ between the human scores ($H$) and the model's scores ($M$) for a given word group and input type is calculated as:

\begin{equation}
r(H, M) = \frac{\sum_{d \in D_{\text{valid}}} (h_d - \mu_H)(m_d - \mu_M)}{\sqrt{\sum_{d \in D_{\text{valid}}} (h_d - \mu_H)^2} \sqrt{\sum_{d \in D_{\text{valid}}} (m_d - \mu_M)^2}}
\end{equation}

\paragraph{Audio Input Type Advantage Score.}

Let $S(g, t, d)$ be the macro F1-score for a given word group $g \in \{\text{natural, constructed}\}$, input type $t \in \{\text{audio, original text}\}$, and semantic dimension $d$.

The Input Type Advantage Score, denoted as $A(g, d)$, for a specific word group $g$ and semantic dimension $d$ is defined as the difference between the scores of the two input types:

\begin{equation}
A(g, d) = S(g, \text{audio}, d) - S(g, \text{original text}, d)
\end{equation}

A positive value of $A(g, d)$ indicates that the ``audio'' input type has an advantage over the ``original text'' input type for that specific dimension and word group. Conversely, a negative value indicates an advantage for the ``original text'' input type.

The score $S(g, t, d)$ is calculated by averaging the macro F1-scores across all models for the given parameters:

\begin{equation}
S(g, t, d) = \frac{1}{|K|} \sum_{k \in K} s_{k,g,t,d}
\end{equation}

Where:
\begin{itemize}
    \item $s_{k,g,t,d}$: the macro F1-score of an individual model $k$.
    \item $K$: the set of all models evaluated.
    \item $|K|$: the total number of models.
\end{itemize}

\subsection{Internal Attention Analysis}

\paragraph{Attention Fraction Score Computation.}
The Phoneme-Semantic Feature Attention Fraction Score is calculated by the algorithm described in Algorithm~\ref{alg:ipa_attn}.

\begin{algorithm*}

\caption{Phoneme-Semantic Feature Attention Fraction Score Calculation.}
\label{alg:ipa_attn}
\begin{algorithmic}[1]

\State \textbf{Load model and prompt}
\State $model \gets \text{Qwen2.5-Omni-7B}$
\State $data \gets$ \Call{load\_data}{}
\State $ipa\_list \gets$ \Call{load\_ipa}{}
\State $dims \gets$ \Call{load\_semantic\_dimensions}{}
\Statex
\State \textbf{Data format:} \{$word$: [($feature_1$, $feature_2$), $answer$]\} \Comment{\textit{feature\_i}: features from semantic dims}

\Statex
\State \textbf{Model Inference Phase}

\ForAll{$word$, ($feature_1$, $feature_2$, $answer$) \textbf{in} data}
    \State $ipa\_text \gets$ \Call{to\_ipa}{$word$} 
    \State $audio \gets$ \Call{to\_audio}{$word$} 
    \ForAll{$input \in \{ipa\_text, audio\}$}
        \State $prompt \gets$ \Call{format\_prompt}{$input$, $feature_1$, $feature_2$} \Comment{Format prompt using input and semantic features}
        \State $response$, $attn$, $tokens \gets$ \Call{inference}{$model$, $prompt$} \Comment{attn = [layer, head, query, key]}
        \State $idx\_input$, $idx\_feature_{1}$, $idx\_feature_{2} \gets$ \Call{find\_idx}{$tokens$, $input$, $feature_1$, $feature_2$} 
        \Comment{type: list[int]}
        \State $idxs \gets \{idx\_input, idx\_feature_{1}, idx\_feature_{2}\}$ \Comment{Aggregate relevant token indices}
        
        \State $filtered\_attn \gets attn[:, :, idxs, idxs]$ 
        \Comment{Select sub-attention map for relevant tokens only}
        \State \Call{save}{$filtered\_attn$, $tokens$, $input$, $idx\_input$, $idx\_feature_{1}$, $idx\_feature_{2}$, $response$}
    \EndFor
\EndFor

\Statex
\State \textbf{Phoneme-Semantic Feature Computation Phase}

\ForAll{$input \in \{ipa\_text, audio\}$}
    \State $result \gets \{\}$

    \ForAll{$word$, ($feature_1$, $feature_2$, $answer$) \textbf{in} data}
        \State $filtered\_attn$, $tokens$, $input$, $idx\_input$, $idx\_feature_{1}$, $idx\_feature_{2}$, $response \gets$ \Call{load}{$word$, $feature_1$, $feature_2$}
        \If{$answer \neq response$}
            \State \textbf{continue}
        \EndIf

        \State $ipa\_span \gets$ \Call{span\_aligning\_input\_to\_tokens}{$input$, $tokens$, $idx\_input$}

        \ForAll{$layers \in filtered\_attn$}
            \ForAll{$layer \in layers$} \Comment{layer = [head, query, key]}
                \ForAll{$idx\_phoneme \in ipa\_span$}
                    \State $score_1 \gets \text{sum}(layer[:, idx\_feature_{1}, idx\_phoneme])$
                    \State $score_2 \gets \text{sum}(layer[:, idx\_feature_{2}, idx\_phoneme])$
                    \State $feature_{1score}$ $\gets$ $score_1 / (score_1 + score_2)$
                    \State $feature_{2score}$ $\gets$ $score_2 / (score_1 + score_2)$

                    \If{$result[phoneme][feature_1][layer]$ is undefined}
                        \State $result[phoneme][feature_1][layer] \gets [\,]$
                    \EndIf
                    \If{$result[phoneme][feature_2][layer]$ is undefined}
                        \State $result[phoneme][feature_2][layer] \gets [\,]$
                    \EndIf

                    \State \Call{append}{$result[phoneme][feature_1][layer]$, $feature_{1score}$}
                    \State \Call{append}{$result[phoneme][feature_2][layer]$, $feature_{2score}$}
                \EndFor
            \EndFor
        \EndFor
    \EndFor

    \ForAll{$phoneme \in ipa\_list$}
        \ForAll{$feature \in dims$}
            \ForAll{$layer \in result[phoneme][feature]$}
                \State $result[phoneme][feature][layer] \gets \text{mean}(result[phoneme][feature][layer])$
            \EndFor
        \EndFor
    \EndFor
    \State \Call{save}{result}
\EndFor

\end{algorithmic}
\end{algorithm*}

\section{Human Evaluation}

\begin{figure*}[htb]
    \centering
    \includegraphics[width=\linewidth]{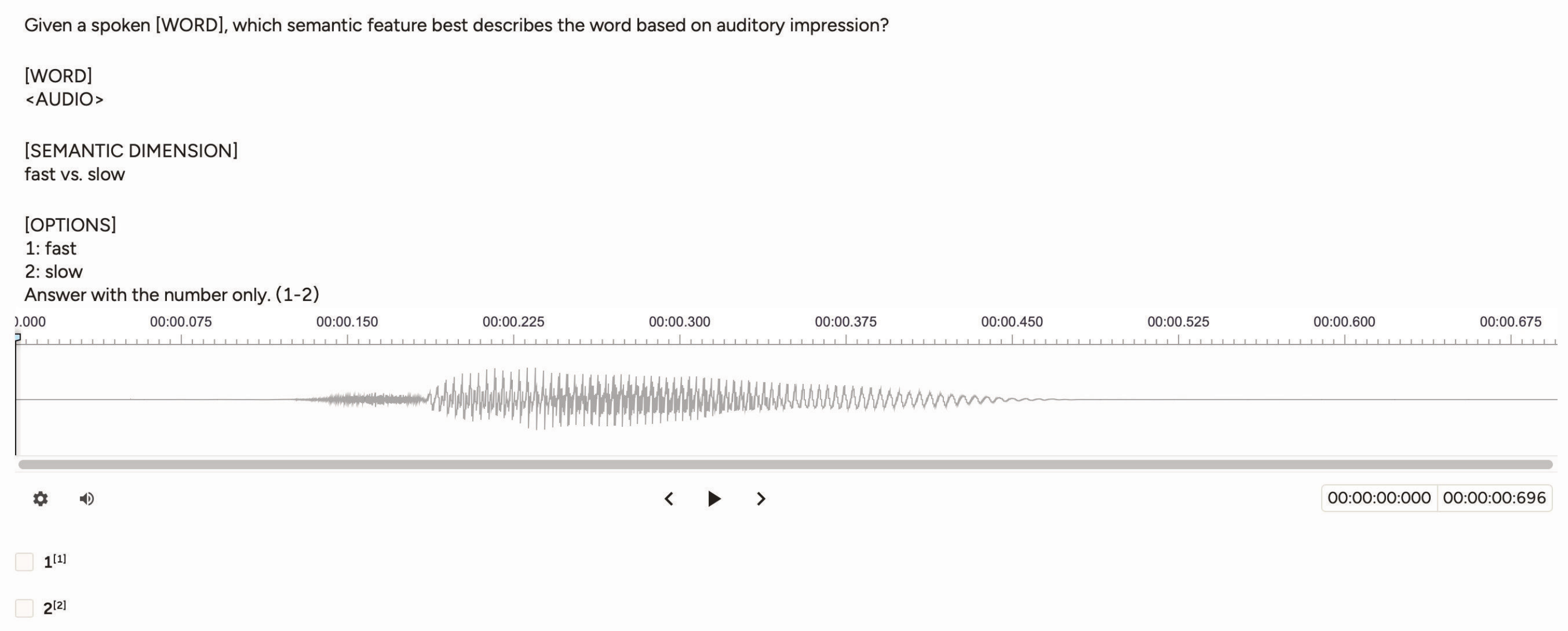}
    \caption{Screenshot image of the human evaluation process. Human evaluators are presented with the same prompts provided to LLMs via the Label Studio interface. Participants select a semantic feature after listening to audio waveform.}
    \label{fig:label_studio_screenshot}
\end{figure*}

To ensure the reliability of the semantic dimension selection and the automatic annotation of the pseudo ground truth data in \dataset, we conduct a human evaluation using 152 questions randomly sampled for each participant from the entire dataset. 10 graduate students without expertise in linguistics participate in this experiment, which is performed via the Label Studio~\citep{LabelStudio}. The experiment utilize questions with the same prompt structure as the semantic dimension prediction test that was provided to LLMs. With regard to the experiment's feasibility, all words are presented in audio waveforms directly hearable to humans. An example image of the process is displayed in Figure~\ref{fig:label_studio_screenshot}.

The randomly sampled test data contains 19 semantic dimensions that are presented in both of the natural and constructed word group data. For each dimension, all natural language types and answer semantic feature distributions are selected from the same distribution, respectively. As a result, four natural group words and four constructed group words are extracted for each dimension.

Due to the nature of sound symbolism task, in which disagreements may arise between responses, the human response distribution may have a wider variance than typical reasoning-centric experiments. However, we maximize reliability within limited resources through the aforementioned methodology. Table~\ref{tab:human_eval_results} shows human response average macro-F1 scores and standard deviation of the scores by semantic dimension.

\begin{table}[htb]
\centering
\begin{tabular}{l c}
\toprule
\textbf{Dimension} & \textbf{Macro-F1 (± Std.)} \\
\midrule
tense-relaxed        & $0.6784\,(\pm\,0.2052)$ \\
realistic-fantastical & $0.6333\,(\pm\,0.2080)$ \\
masculine-feminine   & $0.6287\,(\pm\,0.1643)$ \\
fast-slow            & $0.6210\,(\pm\,0.2185)$ \\
big-small            & $0.6058\,(\pm\,0.1298)$ \\
happy-sad            & $0.6003\,(\pm\,0.1303)$ \\
abrupt-continuous    & $0.5903\,(\pm\,0.2205)$ \\
beautiful-ugly       & $0.5844\,(\pm\,0.1493)$ \\
heavy-light          & $0.5808\,(\pm\,0.1648)$ \\
exciting-calming     & $0.5797\,(\pm\,0.1948)$ \\
sharp-round          & $0.5796\,(\pm\,0.1072)$ \\
hard-soft            & $0.5794\,(\pm\,0.2142)$ \\
solid-nonsolid       & $0.5682\,(\pm\,0.2259)$ \\
inhibited-free       & $0.5621\,(\pm\,0.2263)$ \\
dangerous-safe       & $0.5594\,(\pm\,0.1887)$ \\
harsh-mellow         & $0.5536\,(\pm\,0.1652)$ \\
ordinary-unique      & $0.5376\,(\pm\,0.1891)$ \\
strong-weak          & $0.4863\,(\pm\,0.2000)$ \\
simple-complex       & $0.4602\,(\pm\,0.1752)$ \\
\bottomrule
\end{tabular}
\caption{Human evaluation result with ranked average macro-F1 score and standard deviation by semantic dimension. Among the 25 semantic dimensions, 19 were used for human evaluation due to their presence in both natural and constructed word groups and their sufficient frequency in the dataset.}
\label{tab:human_eval_results}
\end{table}

\section{Detailed Results}

\subsection{Semantic Dimension Prediction Experiment}
\label{appendix:semantic_dimension_prediction_experiment}

Table~\ref{tab:semdim_detailed_Qwen25Omni7B}, \ref{tab:semdim_detailed_Qwen25Omni3B}, \ref{tab:semdim_detailed_gpt4o}, and \ref{tab:semdim_detailed_gemini25flash} show accuracies and macro-F1 scores of experimental models' results for each input types and semantic dimensions. In the tables, ``--'' denotes a dimension where all ground truth features are classified as ``neither'' thus removed.

\begin{table*}[ht]
\centering
\begin{tabular}{l*{12}{c}}
\toprule
\multirow{3}{*}{\textbf{Dimension}} 
& \multicolumn{6}{c}{\textbf{Natural}} & \multicolumn{6}{c}{\textbf{Constructed}} \\
\cmidrule(lr){2-7} \cmidrule(lr){8-13}
& \multicolumn{2}{c}{Original} & \multicolumn{2}{c}{IPA} & \multicolumn{2}{c}{Audio} & \multicolumn{2}{c}{Original} & \multicolumn{2}{c}{IPA} & \multicolumn{2}{c}{Audio} \\
\cmidrule(lr){2-3} \cmidrule(lr){4-5} \cmidrule(lr){6-7} \cmidrule(lr){8-9} \cmidrule(lr){10-11} \cmidrule(lr){12-13}
& Acc. & F1 & Acc. & F1 & Acc. & F1 & Acc. & F1 & Acc. & F1 & Acc. & F1 \\
\midrule
good-bad                            & 67.0 & 61.9 & 38.9 & 38.4 & 53.1 & 48.8 & -- & -- & -- & -- & -- & -- \\ 
beautiful-ugly                      & 68.6 & 62.6 & 28.6 & 25.5 & 48.9 & 43.1 & 72.9 & 60.5 & 77.1 & 54.3 & 68.0 & 50.6 \\ 
pleasant-unpleasant                 & 68.4 & 63.6 & 43.9 & 43.0 & 56.5 & 53.1 & -- & -- & -- & -- & -- & -- \\ 
strong-weak                         & 71.0 & 68.2 & 55.8 & 52.1 & 66.4 & 62.6 & 60.1 & 60.0 & 52.4 & 51.9 & 65.9 & 64.3 \\ 
big-small                           & 61.6 & 54.3 & 62.5 & 60.9 & 60.5 & 54.1 & 49.1 & 39.0 & 64.4 & 64.0 & 50.3 & 46.9 \\ 
rugged-delicate                     & 65.9 & 64.8 & 50.1 & 48.7 & 59.9 & 58.3 & -- & -- & -- & -- & -- & -- \\ 
active-passive                      & 72.4 & 54.3 & 62.8 & 45.8 & 71.0 & 51.9 & -- & -- & -- & -- & -- & -- \\ 
fast-slow                           & 66.7 & 56.9 & 54.6 & 46.0 & 61.0 & 53.4 & 58.5 & 58.5 & 64.4 & 62.1 & 67.7 & 67.6 \\ 
sharp-round                         & 76.6 & 72.1 & 62.7 & 59.2 & 69.8 & 65.4 & 53.7 & 53.7 & 73.1 & 73.0 & 58.4 & 58.0 \\ 
realistic-fantastical               & 68.0 & 39.8 & 19.7 & 15.9 & 75.3 & 43.0 & 60.3 & 53.4 & 58.5 & 51.8 & 79.8 & 79.8 \\ 
structured-disorganized             & 69.8 & 65.0 & 24.7 & 21.5 & 57.4 & 54.8 & -- & -- & -- & -- & -- & -- \\ 
ordinary-unique                     & 76.8 & 50.8 & 74.3 & 48.0 & 68.2 & 48.9 & 51.9 & 35.9 & 64.4 & 58.2 & 63.9 & 61.0 \\ 
interesting-uninteresting           & 72.9 & 63.2 & 66.0 & 38.7 & 67.6 & 45.6 & -- & -- & -- & -- & -- & -- \\ 
simple-complex                      & 69.0 & 44.1 & 70.8 & 49.1 & 68.2 & 41.2 & 49.8 & 34.6 & 49.0 & 33.0 & 48.8 & 33.2 \\ 
abrupt-continuous                   & 75.1 & 71.5 & 58.5 & 54.8 & 66.9 & 64.7 & 46.8 & 42.8 & 61.0 & 54.8 & 53.9 & 53.6 \\ 
exciting-calming                    & 51.3 & 47.3 & 22.1 & 19.9 & 42.9 & 39.9 & 52.6 & 41.3 & 59.9 & 52.0 & 53.6 & 48.4 \\ 
hard-soft                           & 71.7 & 70.9 & 55.2 & 52.8 & 64.7 & 63.1 & 57.4 & 53.5 & 64.8 & 58.7 & 57.2 & 55.8 \\ 
happy-sad                           & 75.8 & 74.4 & 54.4 & 48.4 & 62.2 & 58.2 & 70.3 & 70.0 & 56.7 & 51.1 & 69.0 & 68.1 \\ 
harsh-mellow                        & 84.9 & 68.8 & 55.8 & 48.7 & 78.5 & 61.2 & 54.7 & 51.3 & 67.3 & 66.8 & 51.5 & 46.7 \\ 
heavy-light                         & 74.7 & 73.7 & 56.1 & 55.3 & 66.9 & 65.7 & 64.8 & 64.5 & 50.4 & 47.8 & 46.1 & 43.2 \\ 
inhibited-free                      & 69.1 & 67.3 & 54.0 & 53.4 & 63.7 & 61.2 & 32.0 & 31.8 & 30.6 & 30.2 & 41.7 & 35.9 \\ 
masculine-feminine                  & 65.6 & 61.8 & 39.3 & 36.9 & 57.9 & 55.8 & 62.2 & 62.2 & 66.6 & 66.3 & 68.0 & 68.0 \\ 
solid-nonsolid                      & 75.8 & 69.1 & 63.1 & 55.5 & 70.7 & 61.1 & 87.5 & 52.9 & 77.4 & 56.2 & 81.1 & 60.6 \\ 
tense-relaxed                       & 74.7 & 66.8 & 56.5 & 50.9 & 64.6 & 56.6 & 55.8 & 55.8 & 77.7 & 77.5 & 65.5 & 64.6 \\ 
dangerous-safe                      & 70.5 & 58.7 & 54.8 & 43.0 & 53.6 & 45.3 & 61.6 & 49.4 & 57.9 & 50.0 & 59.0 & 52.7 \\ 
\bottomrule
\end{tabular}
\caption{Detailed semantic dimension prediction accuracy and macro-F1 score results for Qwen2.5-Omni-7B.}
\label{tab:semdim_detailed_Qwen25Omni7B}
\end{table*}

\begin{table*}[ht]
\centering
\begin{tabular}{l*{12}{c}}
\toprule
\multirow{3}{*}{\textbf{Dimension}} 
& \multicolumn{6}{c}{\textbf{Natural}} & \multicolumn{6}{c}{\textbf{Constructed}} \\
\cmidrule(lr){2-7} \cmidrule(lr){8-13}
& \multicolumn{2}{c}{Original} & \multicolumn{2}{c}{IPA} & \multicolumn{2}{c}{Audio} & \multicolumn{2}{c}{Original} & \multicolumn{2}{c}{IPA} & \multicolumn{2}{c}{Audio} \\
\cmidrule(lr){2-3} \cmidrule(lr){4-5} \cmidrule(lr){6-7} \cmidrule(lr){8-9} \cmidrule(lr){10-11} \cmidrule(lr){12-13}
& Acc. & F1 & Acc. & F1 & Acc. & F1 & Acc. & F1 & Acc. & F1 & Acc. & F1 \\
\midrule
good-bad                            & 70.3 & 62.7 & 60.1 & 49.3 & 60.0 & 55.0 & -- & -- & -- & -- & -- & -- \\ 
beautiful-ugly                      & 72.6 & 64.6 & 64.0 & 47.3 & 60.9 & 56.1 & 43.7 & 40.5 & 38.7 & 38.7 & 67.5 & 52.1 \\ 
pleasant-unpleasant                 & 64.0 & 59.7 & 50.3 & 46.2 & 42.5 & 40.7 & -- & -- & -- & -- & -- & -- \\ 
strong-weak                         & 53.6 & 53.4 & 38.4 & 27.4 & 62.8 & 54.2 & 49.0 & 34.5 & 48.6 & 32.7 & 62.0 & 55.6 \\ 
big-small                           & 57.1 & 47.9 & 51.7 & 35.8 & 56.4 & 46.8 & 46.7 & 32.2 & 47.4 & 34.1 & 51.4 & 44.1 \\ 
rugged-delicate                     & 67.9 & 67.2 & 44.5 & 33.9 & 63.4 & 62.2 & -- & -- & -- & -- & -- & -- \\ 
active-passive                      & 49.9 & 42.8 & 14.4 & 12.5 & 76.2 & 53.0 & -- & -- & -- & -- & -- & -- \\ 
fast-slow                           & 51.1 & 47.2 & 26.1 & 23.8 & 53.2 & 47.0 & 58.9 & 55.6 & 49.1 & 39.6 & 61.7 & 61.7 \\ 
sharp-round                         & 77.2 & 69.4 & 63.3 & 59.4 & 80.3 & 70.4 & 55.5 & 53.8 & 73.2 & 72.7 & 54.3 & 47.2 \\ 
realistic-fantastical               & 18.7 & 15.6 & 0.6 & 0.6 & 20.2 & 16.5 & 52.3 & 38.4 & 50.1 & 33.7 & 51.9 & 40.7 \\ 
structured-disorganized             & 70.6 & 62.7 & 44.9 & 43.9 & 51.4 & 50.1 & -- & -- & -- & -- & -- & -- \\ 
ordinary-unique                     & 68.6 & 60.6 & 74.8 & 51.3 & 54.2 & 49.4 & 66.8 & 63.7 & 66.3 & 61.8 & 64.4 & 64.2 \\ 
interesting-uninteresting           & 70.4 & 58.5 & 50.2 & 42.4 & 68.1 & 48.3 & -- & -- & -- & -- & -- & -- \\ 
simple-complex                      & 70.2 & 57.5 & 51.7 & 41.0 & 69.4 & 51.9 & 56.9 & 53.2 & 60.4 & 56.1 & 49.3 & 34.7 \\ 
abrupt-continuous                   & 69.2 & 66.8 & 46.2 & 34.7 & 63.7 & 60.2 & 42.7 & 34.0 & 55.9 & 35.9 & 54.7 & 48.8 \\ 
exciting-calming                    & 58.6 & 51.4 & 23.5 & 20.9 & 47.9 & 43.5 & 71.5 & 71.4 & 62.0 & 56.0 & 62.9 & 61.8 \\ 
hard-soft                           & 57.6 & 50.4 & 48.8 & 32.8 & 54.1 & 44.5 & 61.2 & 38.2 & 61.3 & 38.0 & 61.4 & 39.9 \\ 
happy-sad                           & 68.7 & 66.4 & 55.3 & 52.6 & 66.2 & 60.0 & 68.6 & 68.1 & 39.6 & 35.5 & 52.4 & 44.5 \\ 
harsh-mellow                        & 67.9 & 57.9 & 51.0 & 45.1 & 58.2 & 49.3 & 47.4 & 42.7 & 53.4 & 50.4 & 55.6 & 52.7 \\ 
heavy-light                         & 64.0 & 57.8 & 55.9 & 35.7 & 60.1 & 49.7 & 53.4 & 35.4 & 53.9 & 35.6 & 54.3 & 37.3 \\ 
inhibited-free                      & 61.6 & 54.4 & 53.0 & 34.4 & 56.9 & 43.7 & 47.6 & 33.0 & 49.0 & 32.9 & 48.1 & 32.5 \\ 
masculine-feminine                  & 71.0 & 64.0 & 44.2 & 32.2 & 66.9 & 60.2 & 58.0 & 49.8 & 56.9 & 41.3 & 71.5 & 70.7 \\ 
solid-nonsolid                      & 68.9 & 49.4 & 67.1 & 39.9 & 67.8 & 48.5 & 87.0 & 47.4 & 88.2 & 46.9 & 84.3 & 49.1 \\ 
tense-relaxed                       & 62.2 & 58.0 & 31.1 & 27.0 & 55.1 & 50.5 & 63.1 & 63.1 & 64.1 & 60.0 & 68.9 & 68.9 \\ 
dangerous-safe                      & 43.2 & 40.4 & 15.2 & 13.8 & 32.9 & 30.0 & 19.0 & 16.3 & 19.6 & 17.0 & 25.0 & 23.9 \\ 
\bottomrule
\end{tabular}
\caption{Detailed semantic dimension prediction accuracy and macro-F1 score results for Qwen2.5-Omni-3B.}
\label{tab:semdim_detailed_Qwen25Omni3B}
\end{table*}

\begin{table*}[ht]
\centering
\begin{tabular}{l*{12}{c}}
\toprule
\multirow{3}{*}{\textbf{Dimension}} 
& \multicolumn{6}{c}{\textbf{Natural}} & \multicolumn{6}{c}{\textbf{Constructed}} \\
\cmidrule(lr){2-7} \cmidrule(lr){8-13}
& \multicolumn{2}{c}{Original} & \multicolumn{2}{c}{IPA} & \multicolumn{2}{c}{Audio} & \multicolumn{2}{c}{Original} & \multicolumn{2}{c}{IPA} & \multicolumn{2}{c}{Audio} \\
\cmidrule(lr){2-3} \cmidrule(lr){4-5} \cmidrule(lr){6-7} \cmidrule(lr){8-9} \cmidrule(lr){10-11} \cmidrule(lr){12-13}
& Acc. & F1 & Acc. & F1 & Acc. & F1 & Acc. & F1 & Acc. & F1 & Acc. & F1 \\
\midrule
good-bad                            & 76.1 & 71.4 & 48.9 & 46.8 & 54.2 & 51.3 & -- & -- & -- & -- & -- & -- \\ 
beautiful-ugly                      & 82.7 & 78.0 & 50.3 & 46.3 & 56.7 & 52.4 & 74.5 & 47.2 & 77.3 & 53.9 & 72.1 & 46.0 \\ 
pleasant-unpleasant                 & 75.8 & 71.3 & 50.4 & 48.0 & 51.5 & 49.8 & -- & -- & -- & -- & -- & -- \\ 
strong-weak                         & 78.8 & 77.5 & 65.7 & 61.7 & 64.3 & 63.3 & 77.4 & 77.4 & 67.8 & 67.7 & 77.9 & 77.9 \\ 
big-small                           & 83.4 & 82.1 & 69.4 & 68.4 & 68.8 & 65.9 & 63.0 & 59.3 & 75.3 & 74.8 & 64.5 & 62.2 \\ 
rugged-delicate                     & 80.7 & 79.8 & 65.7 & 64.0 & 67.7 & 65.7 & -- & -- & -- & -- & -- & -- \\ 
active-passive                      & 86.3 & 63.7 & 77.9 & 50.2 & 69.3 & 52.3 & -- & -- & -- & -- & -- & -- \\ 
fast-slow                           & 89.1 & 75.7 & 78.1 & 58.3 & 65.2 & 55.1 & 66.9 & 62.0 & 71.1 & 68.5 & 72.2 & 72.0 \\ 
sharp-round                         & 85.4 & 81.8 & 75.2 & 70.3 & 59.0 & 56.5 & 70.9 & 69.7 & 77.2 & 76.9 & 62.5 & 57.5 \\ 
realistic-fantastical               & 23.9 & 18.8 & 13.4 & 10.7 & 32.5 & 23.6 & 50.5 & 35.3 & 39.3 & 30.7 & 63.1 & 61.9 \\ 
structured-disorganized             & 83.9 & 76.7 & 37.2 & 36.8 & 52.3 & 51.1 & -- & -- & -- & -- & -- & -- \\ 
ordinary-unique                     & 77.4 & 51.0 & 77.6 & 58.4 & 68.3 & 46.8 & 51.0 & 33.8 & 61.1 & 53.3 & 51.9 & 43.1 \\ 
interesting-uninteresting           & 70.9 & 50.6 & 68.8 & 52.7 & 70.4 & 58.5 & -- & -- & -- & -- & -- & -- \\ 
simple-complex                      & 73.1 & 58.2 & 74.0 & 57.3 & 71.5 & 55.5 & 52.7 & 41.0 & 49.9 & 35.1 & 54.6 & 46.3 \\ 
abrupt-continuous                   & 80.5 & 74.6 & 67.3 & 62.8 & 71.5 & 67.6 & 60.3 & 57.9 & 64.8 & 64.2 & 67.5 & 67.3 \\ 
exciting-calming                    & 84.1 & 73.5 & 55.9 & 50.8 & 49.8 & 45.4 & 66.2 & 65.9 & 74.0 & 73.3 & 54.4 & 46.7 \\ 
hard-soft                           & 80.5 & 80.2 & 64.5 & 63.9 & 67.2 & 65.2 & 63.4 & 45.4 & 71.6 & 64.1 & 71.2 & 64.2 \\ 
happy-sad                           & 79.4 & 74.7 & 58.8 & 52.1 & 60.1 & 58.0 & 49.4 & 35.5 & 51.9 & 40.2 & 62.5 & 61.5 \\ 
harsh-mellow                        & 75.5 & 65.5 & 53.1 & 47.0 & 47.8 & 43.5 & 59.8 & 47.6 & 60.6 & 46.4 & 62.2 & 52.7 \\ 
heavy-light                         & 81.6 & 79.6 & 70.6 & 69.3 & 68.1 & 61.7 & 57.4 & 43.6 & 73.9 & 71.0 & 55.6 & 40.8 \\ 
inhibited-free                      & 73.6 & 71.4 & 64.4 & 61.9 & 60.0 & 52.4 & 50.5 & 38.1 & 51.9 & 40.9 & 47.1 & 33.5 \\ 
masculine-feminine                  & 83.5 & 77.4 & 68.3 & 58.4 & 71.2 & 63.4 & 69.4 & 67.0 & 78.1 & 78.0 & 82.7 & 82.2 \\ 
solid-nonsolid                      & 81.0 & 76.8 & 63.0 & 60.8 & 69.5 & 66.3 & 87.9 & 60.2 & 67.7 & 59.3 & 70.8 & 61.4 \\ 
tense-relaxed                       & 79.4 & 73.4 & 66.3 & 57.6 & 68.0 & 61.8 & 71.8 & 71.7 & 68.0 & 66.3 & 85.9 & 85.9 \\ 
dangerous-safe                      & 63.2 & 54.9 & 45.0 & 39.9 & 28.8 & 26.0 & 40.9 & 40.6 & 27.7 & 27.1 & 19.4 & 16.7 \\ 
\bottomrule
\end{tabular}
\caption{Detailed semantic dimension prediction accuracy and macro-F1 score results for gpt-4o.}
\label{tab:semdim_detailed_gpt4o}
\end{table*}

\begin{table*}[ht]
\centering
\begin{tabular}{l*{12}{c}}
\toprule
\multirow{3}{*}{\textbf{Dimension}} 
& \multicolumn{6}{c}{\textbf{Natural}} & \multicolumn{6}{c}{\textbf{Constructed}} \\
\cmidrule(lr){2-7} \cmidrule(lr){8-13}
& \multicolumn{2}{c}{Original} & \multicolumn{2}{c}{IPA} & \multicolumn{2}{c}{Audio} & \multicolumn{2}{c}{Original} & \multicolumn{2}{c}{IPA} & \multicolumn{2}{c}{Audio} \\
\cmidrule(lr){2-3} \cmidrule(lr){4-5} \cmidrule(lr){6-7} \cmidrule(lr){8-9} \cmidrule(lr){10-11} \cmidrule(lr){12-13}
& Acc. & F1 & Acc. & F1 & Acc. & F1 & Acc. & F1 & Acc. & F1 & Acc. & F1 \\
\midrule
good-bad                            & 86.0 & 80.2 & 76.4 & 57.7 & 78.0 & 53.5 & -- & -- & -- & -- & -- & -- \\ 
beautiful-ugly                      & 90.7 & 86.7 & 72.2 & 60.1 & 62.4 & 55.7 & 72.7 & 65.1 & 67.1 & 64.3 & 39.4 & 38.0 \\ 
pleasant-unpleasant                 & 82.2 & 76.8 & 71.4 & 59.7 & 79.2 & 52.3 & -- & -- & -- & -- & -- & -- \\ 
strong-weak                         & 76.7 & 73.8 & 60.5 & 58.9 & 62.9 & 56.3 & 62.0 & 61.5 & 66.8 & 66.0 & 47.6 & 45.7 \\ 
big-small                           & 83.3 & 82.8 & 65.0 & 61.0 & 59.6 & 55.2 & 63.0 & 59.4 & 62.5 & 59.0 & 50.5 & 50.0 \\ 
rugged-delicate                     & 83.2 & 82.3 & 68.0 & 65.9 & 66.0 & 56.7 & -- & -- & -- & -- & -- & -- \\ 
active-passive                      & 87.5 & 59.8 & 70.2 & 49.6 & 68.2 & 49.6 & -- & -- & -- & -- & -- & -- \\ 
fast-slow                           & 80.7 & 70.0 & 47.6 & 44.1 & 30.9 & 30.5 & 63.9 & 63.9 & 58.0 & 51.9 & 50.9 & 41.5 \\ 
sharp-round                         & 86.5 & 82.2 & 73.3 & 68.8 & 76.9 & 65.3 & 61.2 & 60.2 & 73.1 & 71.9 & 56.6 & 52.1 \\ 
realistic-fantastical               & 69.6 & 40.9 & 33.5 & 24.7 & 86.4 & 46.4 & 51.1 & 48.4 & 67.3 & 66.3 & 55.7 & 53.1 \\ 
structured-disorganized             & 84.2 & 78.2 & 67.0 & 58.0 & 69.4 & 56.8 & -- & -- & -- & -- & -- & -- \\ 
ordinary-unique                     & 80.7 & 72.3 & 68.6 & 56.3 & 51.3 & 46.4 & 54.8 & 46.0 & 62.0 & 61.5 & 56.7 & 55.9 \\ 
interesting-uninteresting           & 74.2 & 59.9 & 62.7 & 47.7 & 62.3 & 51.0 & -- & -- & -- & -- & -- & -- \\ 
simple-complex                      & 80.8 & 68.4 & 67.8 & 57.2 & 54.6 & 47.9 & 57.9 & 51.6 & 58.7 & 57.0 & 49.9 & 49.9 \\ 
abrupt-continuous                   & 84.2 & 81.3 & 73.0 & 69.6 & 71.7 & 61.5 & 55.4 & 55.3 & 61.7 & 57.5 & 50.0 & 48.3 \\ 
exciting-calming                    & 84.5 & 71.8 & 65.6 & 54.1 & 58.2 & 50.7 & 54.8 & 50.1 & 71.8 & 71.7 & 56.2 & 55.7 \\ 
hard-soft                           & 79.8 & 79.6 & 64.2 & 62.7 & 59.6 & 58.3 & 63.6 & 58.2 & 67.3 & 56.6 & 53.1 & 51.9 \\ 
happy-sad                           & 81.3 & 80.3 & 57.2 & 55.7 & 51.0 & 42.2 & 59.7 & 59.5 & 60.7 & 58.8 & 53.0 & 39.2 \\ 
harsh-mellow                        & 85.6 & 73.8 & 70.1 & 57.1 & 75.5 & 56.1 & 59.9 & 57.6 & 60.7 & 51.9 & 49.5 & 46.7 \\ 
heavy-light                         & 82.2 & 81.7 & 70.0 & 69.4 & 56.4 & 48.8 & 66.8 & 61.3 & 75.3 & 73.3 & 47.1 & 35.0 \\ 
inhibited-free                      & 74.4 & 73.9 & 60.9 & 60.1 & 55.8 & 51.6 & 46.6 & 44.7 & 47.1 & 44.4 & 50.0 & 40.9 \\ 
masculine-feminine                  & 82.5 & 76.2 & 59.6 & 52.7 & 54.0 & 51.7 & 67.0 & 61.8 & 65.0 & 60.5 & 58.1 & 58.0 \\ 
solid-nonsolid                      & 84.8 & 82.0 & 68.0 & 60.8 & 47.5 & 46.2 & 88.2 & 61.1 & 87.9 & 73.4 & 25.6 & 25.6 \\ 
tense-relaxed                       & 76.3 & 71.4 & 64.0 & 55.8 & 69.4 & 58.5 & 61.7 & 61.0 & 79.6 & 79.5 & 65.5 & 62.2 \\ 
dangerous-safe                      & 78.8 & 65.3 & 73.7 & 52.3 & 80.3 & 52.2 & 66.9 & 56.4 & 53.6 & 51.3 & 77.3 & 46.6 \\ 
\bottomrule
\end{tabular}
\caption{Detailed semantic dimension prediction accuracy and macro-F1 score results for gemini-2.5-flash.}
\label{tab:semdim_detailed_gemini25flash}
\end{table*}

\subsection{Internal Attention Analysis}

We provide full result tables for the Phoneme-Semantic Feature Attention Fraction Scores by word group (natural and constructed) and input type (IPA text and audio waveform).

\begin{itemize}
    \item Natural - IPA text: Table~\ref{tab:natural_ipa_text_ipa_semdim_attention_part1}, \ref{tab:natural_ipa_text_ipa_semdim_attention_part2}, \ref{tab:natural_ipa_text_ipa_semdim_attention_part3}, and \ref{tab:natural_ipa_text_ipa_semdim_attention_part4}.
    \item Natural - Audio waveform: Table~\ref{tab:natural_audio_ipa_semdim_attention_part1}, \ref{tab:natural_audio_ipa_semdim_attention_part2}, \ref{tab:natural_audio_ipa_semdim_attention_part3}, and \ref{tab:natural_audio_ipa_semdim_attention_part4}.
    \item Constructed - IPA text: Table~\ref{tab:constructed_ipa_text_ipa_semdim_attention_part1} and \ref{tab:constructed_ipa_text_ipa_semdim_attention_part2}.
    \item Constructed - Audio waveform: Table~\ref{tab:constructed_audio_ipa_semdim_attention_part1} and \ref{tab:constructed_audio_ipa_semdim_attention_part2}.
\end{itemize}

\begin{table*}[htb]
\centering
\begin{tabular}{lcccccccccccccc}
\toprule
\textbf{Dimension} & \textipa{i} & \textipa{I} & \textipa{e} & \textipa{E} & \textipa{\ae} & \textipa{a} & \textipa{W} & \textipa{u} & \textipa{U} & \textipa{o} & \textipa{2} & \textipa{O} & \textipa{A} & \textipa{@} \\
\midrule
abrupt & .506 & .491 & .512 & .511 & .510 & .517 & .503 & .506 & .424 & .509 & .510 & .492 & .506 & .511 \\
continuous & .494 & .509 & .488 & .489 & .490 & .482 & .497 & .494 & .576 & .490 & .490 & .508 & .493 & .489 \\
\midrule
active & .495 & .499 & .493 & .482 & .507 & .502 & .469 & .496 & .407 & .496 & .493 & .461 & .505 & .492 \\
passive & .505 & .501 & .507 & .517 & .493 & .498 & .530 & .504 & .593 & .504 & .507 & .539 & .495 & .508 \\
\midrule
beautiful & .496 & .455 & .462 & .499 & .425 & .487 & .470 & .443 & .506 & .484 & .426 & .442 & .405 & .486 \\
ugly & .504 & .545 & .538 & .501 & .575 & .513 & .529 & .556 & .494 & .516 & .573 & .558 & .594 & .514 \\
\midrule
big & .492 & .473 & .551 & .452 & .495 & .469 & .497 & .515 & .501 & .465 & .559 & .531 & .559 & .506 \\
\bottomrule
\end{tabular}
\caption{Phoneme-Semantic Feature Attention Fraction Scores (Natural words - IPA text - Part 1).}
\label{tab:natural_ipa_text_ipa_semdim_attention_part1}
\end{table*}

\begin{table*}[htb]
\centering
\begin{tabular}{lcccccccccccccc}
\toprule
\textbf{Dimension} & \textipa{\~W} & \textipa{W:} & \textipa{a:} & \textipa{o:} & \textipa{i:} & \textipa{e:} & \textipa{b} & \textipa{d} & \textipa{g} & \textipa{p} & \textipa{t} & \textipa{k} & \textipa{v} & \textipa{D} \\
\midrule
abrupt & .500 & .502 & .508 & .504 & .504 & -- & .523 & .517 & .501 & .519 & .512 & .515 & .500 & .534 \\
continuous & .500 & .498 & .492 & .496 & .496 & -- & .477 & .482 & .499 & .481 & .488 & .485 & .499 & .466 \\
\midrule
active & .423 & .465 & .509 & -- & .492 & .597 & .497 & .504 & .491 & .495 & .498 & .498 & .482 & -- \\
passive & .577 & .535 & .491 & -- & .508 & .403 & .503 & .496 & .508 & .505 & .502 & .502 & .517 & -- \\
\midrule
beautiful & -- & .564 & .432 & -- & .435 & .404 & .438 & .419 & .416 & .483 & .459 & .444 & .388 & -- \\
ugly & -- & .435 & .568 & -- & .565 & .596 & .561 & .581 & .584 & .517 & .540 & .555 & .612 & -- \\
\midrule
big & -- & .636 & .596 & .613 & .533 & -- & .547 & .567 & .570 & .507 & .501 & .500 & .635 & -- \\
\bottomrule
\end{tabular}
\caption{Phoneme-Semantic Feature Attention Fraction Scores (Natural words - IPA text - Part 2).}
\label{tab:natural_ipa_text_ipa_semdim_attention_part2}
\end{table*}

\begin{table*}[htb]
\centering
\begin{tabular}{lcccccccccccccc}
\toprule
\textbf{Dimension} & \textipa{z} & \textipa{Z} & \textipa{f} & \textipa{s} & \textipa{S} & \textipa{T} & \textipa{\c{c}} & \textipa{h} & \textipa{F} & \textipa{m} & \textipa{n} & \textipa{\textltailn} & \textipa{N} & N \\
\midrule
abrupt & .515 & .488 & .528 & .508 & .516 & .510 & .505 & .516 & .515 & .511 & .511 & .504 & .500 & .474 \\
continuous & .485 & .512 & .472 & .491 & .484 & .490 & .494 & .484 & .485 & .489 & .489 & .496 & .500 & .526 \\
\midrule
active & .503 & .495 & .510 & .494 & .505 & .513 & .496 & .507 & .495 & .497 & .488 & .483 & .493 & .468 \\
passive & .497 & .505 & .490 & .506 & .495 & .487 & .504 & .492 & .504 & .502 & .511 & .517 & .507 & .532 \\
\midrule
beautiful & .451 & .396 & .435 & .490 & .451 & .409 & .491 & .474 & .418 & .455 & .453 & .485 & .487 & .501 \\
ugly & .548 & .604 & .565 & .510 & .549 & .591 & .509 & .526 & .582 & .545 & .547 & .515 & .513 & .499 \\
\midrule
big & .507 & .559 & .499 & .473 & .561 & .617 & .459 & .510 & .491 & .497 & .486 & -- & .507 & .535 \\
\bottomrule
\end{tabular}
\caption{Phoneme-Semantic Feature Attention Fraction Scores (Natural words - IPA text - Part 3).}
\label{tab:natural_ipa_text_ipa_semdim_attention_part3}
\end{table*}

\begin{table*}[htb]
\centering
\begin{tabular}{lccccccccccccc}
\toprule
\textbf{Dimension} & \textipa{l} & R & \textipa{R} & \textipa{\*r} & \textipa{j} & \textipa{\textturnmrleg} & \textipa{C} & \textipa{\t{d\textctz}} & \textipa{\t{tC}} & \textipa{\textctz} & \textipa{P} & \textipa{w} & \textipa{4} \\
\midrule
abrupt & .507 & .509 & .506 & .502 & .518 & .493 & .514 & .515 & .511 & .500 & .534 & .507 & .511 \\
continuous & .492 & .490 & .494 & .497 & .482 & .507 & .485 & .485 & .489 & .500 & .466 & .492 & .489 \\
\midrule
active & .491 & .478 & .498 & .495 & .512 & .511 & .508 & .492 & .503 & .501 & .475 & .498 & .471 \\
passive & .508 & .521 & .501 & .505 & .488 & .489 & .492 & .508 & .497 & .499 & .525 & .501 & .529 \\
\midrule
beautiful & .463 & .432 & .496 & .448 & .460 & .514 & .447 & .368 & .441 & .425 & .455 & .443 & .411 \\
ugly & .537 & .568 & .504 & .551 & .540 & .486 & .552 & .632 & .559 & .574 & .545 & .557 & .589 \\
\midrule
big & .484 & .554 & .508 & .506 & .468 & .587 & .505 & .527 & .503 & .480 & .520 & .537 & .491 \\
\bottomrule
\end{tabular}
\caption{Phoneme-Semantic Feature Attention Fraction Scores (Natural words - IPA text - Part 4).}
\label{tab:natural_ipa_text_ipa_semdim_attention_part4}
\end{table*}

\begin{table*}[htb]
\centering
\begin{tabular}{lcccccccccccccc}
\toprule
\textbf{Dimension} & \textipa{i} & \textipa{I} & \textipa{e} & \textipa{E} & \textipa{a} & \textipa{1} & \textipa{W} & \textipa{u} & \textipa{U} & \textipa{o} & \textipa{\textrhookschwa} & \textipa{2} & \textipa{O} & \textipa{A} \\
\midrule
abrupt & .517 & .525 & .526 & .522 & .537 & .518 & .513 & .513 & .557 & .516 & .534 & .524 & .520 & .532 \\
continuous & .483 & .475 & .474 & .478 & .463 & .482 & .486 & .487 & .443 & .484 & .466 & .476 & .480 & .468 \\
\midrule
active & .500 & .506 & .496 & .496 & .502 & .494 & .504 & .496 & .485 & .497 & .509 & .490 & .511 & .511 \\
passive & .500 & .494 & .504 & .504 & .498 & .506 & .496 & .504 & .515 & .503 & .491 & .510 & .489 & .489 \\
\midrule
beautiful & .467 & .467 & .460 & .488 & .479 & .469 & .479 & .439 & .531 & .470 & .479 & .437 & .475 & .466 \\
ugly & .533 & .533 & .540 & .512 & .521 & .531 & .521 & .561 & .469 & .530 & .521 & .563 & .525 & .534 \\
\midrule
big & .496 & .511 & .513 & .488 & .505 & .494 & .498 & .496 & .525 & .484 & .516 & .508 & .517 & .509 \\
\bottomrule
\end{tabular}
\caption{Phoneme-Semantic Feature Attention Fraction Scores (Natural words - Audio - Part 1).}
\label{tab:natural_audio_ipa_semdim_attention_part1}
\end{table*}

\begin{table*}[htb]
\centering
\begin{tabular}{lcccccccccccccc}
\toprule
\textbf{Dimension} & \textipa{5} & \textipa{@} & \textipa{W:} & \textipa{a:} & \textipa{o:} & \textipa{i:} & \textipa{b} & \textipa{d} & \textipa{g} & \textipa{p} & \textipa{t} & \textipa{c} & \textipa{k} & \textipa{v} \\
\midrule
abrupt & .530 & .536 & .515 & .541 & .546 & .555 & .516 & .518 & .515 & .518 & .524 & .514 & .521 & .505 \\
continuous & .470 & .464 & .485 & .459 & .454 & .445 & .484 & .482 & .485 & .482 & .476 & .486 & .479 & .495 \\
\midrule
active & .494 & .503 & .508 & .516 & .508 & .504 & .507 & .512 & .506 & .504 & .497 & .508 & .502 & .509 \\
passive & .506 & .497 & .492 & .484 & .492 & .496 & .493 & .488 & .494 & .496 & .503 & .492 & .498 & .491 \\
\midrule
beautiful & .464 & .471 & .488 & -- & -- & .444 & .469 & .455 & .479 & .491 & .469 & .505 & .463 & .430 \\
ugly & .536 & .529 & .512 & -- & -- & .556 & .531 & .545 & .521 & .509 & .531 & .495 & .537 & .570 \\
\midrule
big & .486 & .505 & -- & .468 & .513 & -- & .512 & .520 & .505 & .504 & .501 & .494 & .498 & .501 \\
\bottomrule
\end{tabular}
\caption{Phoneme-Semantic Feature Attention Fraction Scores (Natural words - Audio - Part 2).}
\label{tab:natural_audio_ipa_semdim_attention_part2}
\end{table*}

\begin{table*}[htb]
\centering
\begin{tabular}{lcccccccccccccc}
\toprule
\textbf{Dimension} & \textipa{z} & \textipa{K} & \textipa{f} & \textipa{s} & \textipa{S} & \textipa{T} & \textipa{h} & \textipa{F} & \textipa{m} & \textipa{n} & \textipa{\textltailn} & \textipa{N} & N & \textipa{l} \\
\midrule
abrupt & .514 & .534 & .514 & .505 & .525 & .521 & .514 & .513 & .511 & .512 & .500 & .519 & .522 & .518 \\
continuous & .485 & .466 & .486 & .495 & .475 & .479 & .486 & .487 & .489 & .488 & .500 & .481 & .477 & .482 \\
\midrule
active & .508 & .502 & .506 & .499 & .508 & .516 & .505 & .503 & .505 & .505 & .502 & .500 & .520 & .505 \\
passive & .492 & .498 & .494 & .501 & .492 & .484 & .495 & .497 & .495 & .495 & .498 & .500 & .480 & .495 \\
\midrule
beautiful & .464 & .442 & .498 & .487 & .495 & .474 & .488 & .529 & .472 & .473 & .491 & .482 & .505 & .484 \\
ugly & .535 & .558 & .502 & .513 & .504 & .526 & .512 & .471 & .528 & .527 & .509 & .518 & .495 & .516 \\
\midrule
big & .484 & .506 & .509 & .484 & .512 & .478 & .504 & .498 & .492 & .501 & -- & .495 & .500 & .506 \\
\bottomrule
\end{tabular}
\caption{Phoneme-Semantic Feature Attention Fraction Scores (Natural words - Audio - Part 3).}
\label{tab:natural_audio_ipa_semdim_attention_part3}
\end{table*}

\begin{table*}[htb]
\centering
\begin{tabular}{lccccccccc}
\toprule
\textbf{Dimension} & \textipa{l} (retroflex) & \textipa{r} & \textipa{R} & \textipa{j} & \textipa{C} & \textipa{\textctz} & \textipa{P} & \textipa{w} & \textipa{4} \\
\midrule
abrupt & .523 & .525 & .513 & .516 & .513 & .514 & .518 & .523 & .528 \\
continuous & .477 & .475 & .487 & .484 & .487 & .486 & .482 & .477 & .472 \\
\midrule
active & .498 & .508 & .497 & .504 & .511 & .510 & .514 & .504 & .491 \\
passive & .502 & .492 & .503 & .496 & .489 & .490 & .486 & .496 & .509 \\
\midrule
beautiful & .462 & .483 & .488 & .490 & .468 & .449 & .499 & .466 & .431 \\
ugly & .538 & .517 & .512 & .510 & .532 & .551 & .501 & .534 & .569 \\
\midrule
big & .496 & .519 & .479 & .495 & .498 & .492 & .505 & .500 & .501 \\
\bottomrule
\end{tabular}
\caption{Phoneme-Semantic Feature Attention Fraction Scores (Natural words - Audio - Part 4).}
\label{tab:natural_audio_ipa_semdim_attention_part4}
\end{table*}

\begin{table*}[htb]
\centering
\begin{tabular}{lcccccccccccccc}
\toprule
\textbf{Dimension} & \textipa{i} & \textipa{ej} & \textipa{A} & \textipa{ow} & \textipa{b} & \textipa{d} & \textipa{g} & \textipa{p} & \textipa{t} & \textipa{k} & \textipa{v} & \textipa{D} & \textipa{z} & \textipa{f} \\
\midrule
abrupt & .513 & .527 & .514 & .513 & .527 & .533 & .527 & .506 & .497 & .511 & .465 & .460 & .491 & .521 \\
continuous & .487 & .473 & .486 & .486 & .473 & .467 & .472 & .494 & .503 & .489 & .535 & .540 & .509 & .479 \\
\midrule
active & .498 & .498 & .485 & .487 & .498 & .509 & .497 & .495 & .500 & .506 & .504 & .496 & .502 & .498 \\
passive & .501 & .501 & .514 & .512 & .501 & .491 & .502 & .504 & .500 & .494 & .496 & .504 & .497 & .502 \\
\midrule
beautiful & .522 & .534 & .529 & .511 & .434 & .410 & .408 & -- & -- & -- & -- & -- & -- & .543 \\
ugly & .478 & .465 & .471 & .489 & .565 & .590 & .592 & -- & -- & -- & -- & -- & -- & .456 \\
\midrule
big & .463 & .439 & .586 & .567 & .550 & .555 & .547 & .521 & .508 & .505 & .556 & .527 & .541 & .491 \\
\bottomrule
\end{tabular}
\caption{Phoneme-Semantic Feature Attention Fraction Scores (Constructed words - IPA text - Part 1).}
\label{tab:constructed_ipa_text_ipa_semdim_attention_part1}
\end{table*}

\begin{table*}[htb]
\centering
\begin{tabular}{lccccc}
\toprule
\textbf{Dimension} & \textipa{s} & \textipa{S} & \textipa{m} & \textipa{n} & \textipa{l} \\
\midrule
abrupt & .516 & .513 & .515 & .529 & .515 \\
continuous & .484 & .487 & .485 & .471 & .485 \\
\midrule
active & .494 & .498 & .481 & .500 & .489 \\
passive & .505 & .502 & .519 & .500 & .511 \\
\midrule
beautiful & .556 & .529 & .551 & .553 & .537 \\
ugly & .444 & .471 & .449 & .447 & .463 \\
\midrule
big & .478 & .481 & .463 & .465 & .462 \\
\bottomrule
\end{tabular}
\caption{Phoneme-Semantic Feature Attention Fraction Scores (Constructed words - IPA text - Part 2).}
\label{tab:constructed_ipa_text_ipa_semdim_attention_part2}
\end{table*}

\begin{table*}[htb]
\centering
\begin{tabular}{lcccccccccccccc}
\toprule
\textbf{Dimension} & \textipa{i} & \textipa{A} & \textipa{b} & \textipa{d} & \textipa{g} & \textipa{p} & \textipa{t} & \textipa{k} & \textipa{v} & \textipa{z} & \textipa{f} & \textipa{s} & \textipa{S} & \textipa{m} \\
\midrule
abrupt & .522 & .539 & .531 & .521 & .529 & .505 & .513 & .521 & .512 & .520 & .526 & .523 & .529 & .526 \\
continuous & .478 & .461 & .469 & .479 & .471 & .495 & .486 & .479 & .488 & .480 & .474 & .477 & .471 & .474 \\
\midrule
active & .507 & .510 & .516 & .511 & .509 & .515 & .509 & .510 & .513 & .506 & .506 & .513 & .508 & .515 \\
passive & .493 & .490 & .484 & .489 & .491 & .485 & .491 & .490 & .487 & .494 & .494 & .487 & .491 & .485 \\
\midrule
beautiful & .511 & .505 & .477 & .474 & .477 & -- & -- & -- & -- & -- & .522 & .523 & .518 & .517 \\
ugly & .489 & .495 & .523 & .526 & .523 & -- & -- & -- & -- & -- & .478 & .477 & .482 & .483 \\
\midrule
big & .499 & .536 & .521 & .514 & .517 & .502 & .501 & .498 & .516 & .507 & .503 & .493 & .499 & .498 \\
\bottomrule
\end{tabular}
\caption{Phoneme-Semantic Feature Attention Fraction Scores (Constructed words - Audio - Part 1).}
\label{tab:constructed_audio_ipa_semdim_attention_part1}
\end{table*}

\begin{table*}[htb]
\centering
\begin{tabular}{lcc}
\toprule
\textbf{Dimension} & \textipa{n} & \textipa{l} \\
\midrule
abrupt & .522 & .524 \\
continuous & .478 & .476 \\
\midrule
active & .513 & .518 \\
passive & .487 & .482 \\
\midrule
beautiful & .530 & .514 \\
ugly & .470 & .486 \\
\midrule
big & .499 & .501 \\
small & .501 & .499 \\
\midrule
dangerous & .558 & .535 \\
safe & .442 & .465 \\
\midrule
exciting & .503 & .499 \\
calming & .497 & .500 \\
\midrule
fast & .456 & .462 \\
slow & .544 & .538 \\
\midrule
happy & .496 & .495 \\
sad & .504 & .505 \\
\midrule
hard & .482 & .479 \\
soft & .518 & .521 \\
\midrule
harsh & .259 & .272 \\
mellow & .741 & .728 \\
\midrule
heavy & .463 & .462 \\
light & .537 & .538 \\
\midrule
inhibited & .444 & .456 \\
free & .556 & .544 \\
\midrule
masculine & .495 & .496 \\
feminine & .505 & .504 \\
\midrule
realistic & .367 & .356 \\
fantastical & .633 & .644 \\
\midrule
rugged & .498 & .487 \\
delicate & .502 & .513 \\
\midrule
sharp & .489 & .497 \\
round & .511 & .503 \\
\midrule
simple & .519 & .521 \\
complex & .481 & .479 \\
\midrule
solid & .349 & .354 \\
nonsolid & .651 & .646 \\
\midrule
strong & .491 & .492 \\
weak & .509 & .508 \\
\midrule
tense & .511 & .493 \\
relaxed & .489 & .507 \\
\bottomrule
\end{tabular}
\caption{Phoneme-Semantic Feature Attention Fraction Scores (Constructed words - Audio - Part 2).}
\label{tab:constructed_audio_ipa_semdim_attention_part2}
\end{table*}

We also calculate the average attention fraction scores by layer in Qwen2.5-Omni-7B as shown in Table~\ref{tab:layerwise_attn_fraction_score}.

\begin{table*}[htb]
\centering
\begin{tabular}{lcccc}
\toprule
\multirow{2}{*}{\textbf{Layer}} & \multicolumn{2}{c}{\textbf{Natural}} & \multicolumn{2}{c}{\textbf{Constructed}} \\
\cmidrule(lr){2-3} \cmidrule(lr){4-5}
& \textbf{IPA} & \textbf{Audio} & \textbf{IPA} & \textbf{Audio} \\
\midrule
0  & 0.5022 & 0.4974 & 0.5151 & 0.4950 \\
1  & 0.5047 & 0.5000 & 0.5048 & 0.4901 \\
2  & 0.5060 & 0.5023 & 0.5109 & 0.5000 \\
3  & 0.5054 & 0.5002 & 0.5147 & 0.4982 \\
4  & 0.5020 & 0.5036 & 0.5100 & 0.5036 \\
5  & 0.5090 & 0.5018 & 0.5228 & 0.5052 \\
6  & 0.5026 & 0.4985 & 0.5022 & 0.4845 \\
7  & 0.5050 & 0.5007 & 0.5394 & 0.5152 \\
8  & 0.5087 & 0.5049 & 0.5395 & 0.5185 \\
9  & 0.5121 & 0.5076 & 0.5110 & 0.5045 \\
10 & 0.5112 & 0.5030 & 0.5253 & 0.5083 \\
11 & 0.5096 & 0.5022 & 0.5311 & 0.5088 \\
12 & 0.5064 & 0.4957 & 0.5409 & 0.5036 \\
13 & 0.5061 & 0.4964 & 0.5270 & 0.5049 \\
14 & 0.5065 & 0.4962 & 0.5269 & 0.5111 \\
15 & 0.5077 & 0.5046 & 0.5282 & 0.5048 \\
16 & 0.5140 & 0.5022 & 0.5249 & 0.4974 \\
17 & 0.5075 & 0.5011 & 0.5180 & 0.5023 \\
18 & 0.5065 & 0.5058 & 0.5291 & 0.5144 \\
19 & 0.5092 & 0.5002 & 0.5062 & 0.5019 \\
20 & 0.5055 & 0.5056 & 0.5165 & 0.5104 \\
21 & 0.5091 & 0.5016 & 0.5186 & 0.5092 \\
22 & 0.5119 & 0.5002 & 0.5285 & 0.5095 \\
23 & 0.5082 & 0.4985 & 0.5229 & 0.5148 \\
24 & 0.5049 & 0.4991 & 0.5427 & 0.5116 \\
25 & 0.5077 & 0.4998 & 0.5337 & 0.5101 \\
26 & 0.5104 & 0.4983 & 0.5324 & 0.5076 \\
27 & 0.5056 & 0.4987 & 0.5101 & 0.5043 \\
\midrule
\textbf{Average} & \textbf{0.5073} & \textbf{0.5009} & \textbf{0.5226} & \textbf{0.5053} \\
\bottomrule
\end{tabular}
\caption{Layer-wise attention fraction scores for IPA and audio input types across natural and constructed word groups in Qwen2.5-Omni-7B.}
\label{tab:layerwise_attn_fraction_score}
\end{table*}

\section{Prompts}
\label{appendix:prompts}

Table~\ref{tab:prompt_semantic_dimension_annotation} shows an automatic semantic dimension annotation prompt for LLMs. Table~\ref{tab:prompt_semantic_dimension_prediction} represents detailed prompts for the semantic dimension experiments.

\begin{table*}[h]
\centering

\begin{subfigure}{\textwidth}
    \centering
        \begin{tabular}{p{0.9\textwidth}}
            \toprule
                You are a professional linguistic annotator.\\
                Please read a \verb|{language}| mimetic word and its meaning, and decide which semantic feature best describes the word's meaning.\\
                \\
                \textnormal{[}WORD]\\
                \verb|{word}|\\
                \\
                \textnormal{[}MEANING]\\
                \verb|{meaning}|\\
                \\
                \textnormal{[}SEMANTIC DIMENSION]\\
                \verb|{feature1}| vs. \verb|{feature2}|\\
                \\
                \textnormal{[}OPTIONS]\\
                1: \verb|{feature1}|\\
                2: \verb|{feature2}|\\
                3: Neither\\
                Answer with the number only. (1-3)\\
            \bottomrule
        \end{tabular}
\end{subfigure}

\vspace{0.5cm} 

\caption{Prompt for automatic semantic dimension annotation.}
\label{tab:prompt_semantic_dimension_annotation}
\end{table*}

\begin{table*}[htb]
\centering

\begin{subfigure}{\textwidth}
    \centering
        \begin{tabular}{p{0.9\textwidth}}
            \toprule
                Given a [WORD], which semantic feature best describes the word based on auditory impression?\\
                \\
                \textnormal{[}WORD]\\
                \verb|{word}|\\
                \\
                \textnormal{[}SEMANTIC DIMENSION]
                \\
                \verb|{feature1}| vs. \verb|{feature2}|\\
                \\
                \textnormal{[}OPTIONS]\\
                1: \verb|{feature1}|\\
                2: \verb|{feature2}|\\
                Answer with the number only. (1-2)\\
            \bottomrule
        \end{tabular}
    \caption{Prompt for original text words.}
\end{subfigure}

\vspace{0.5cm} 

\begin{subfigure}{\textwidth}
    \centering
        \begin{tabular}{p{0.9\textwidth}}
            \toprule
                Given an IPA [WORD], which semantic feature best describes the word based on auditory impression?\\
                \\
                \textnormal{[}WORD]\\
                \verb|{word}|\\
                \\
                \textnormal{[}SEMANTIC DIMENSION]
                \\
                \verb|{feature1}| vs. \verb|{feature2}|\\
                \\
                \textnormal{[}OPTIONS]\\
                1: \verb|{feature1}|\\
                2: \verb|{feature2}|\\
                Answer with the number only. (1-2)\\
            \bottomrule
        \end{tabular}
    \caption{Prompt for IPA text words.}
\end{subfigure}

\vspace{0.5cm} 

\begin{subfigure}{\textwidth}
    \centering
        \begin{tabular}{p{0.9\textwidth}}
            \toprule
                Given a spoken [WORD], which semantic feature best describes the word based on auditory impression?\\
                \\
                \textnormal{[}WORD]\\
                \verb|{audio}|\\
                \\
                \textnormal{[}SEMANTIC DIMENSION]
                \\
                \verb|{feature1}| vs. \verb|{feature2}|\\
                \\
                \textnormal{[}OPTIONS]\\
                1: \verb|{feature1}|\\
                2: \verb|{feature2}|\\
                Answer with the number only. (1-2)\\
            \bottomrule
        \end{tabular}
    \caption{Prompt for spoken audio words.}
\end{subfigure}

\caption{Prompts for the semantic dimension prediction experiment by input type.}
\label{tab:prompt_semantic_dimension_prediction}
\end{table*}

\end{document}